\documentclass[10pt,twocolumn,letterpaper]{article}
\usepackage[misc,geometry]{ifsym}
\usepackage{iccv}
\usepackage{times}
\usepackage{epsfig}
\usepackage{graphicx}
\usepackage{amsmath}
\usepackage{amssymb}
\usepackage{pifont}
\usepackage{caption}
\usepackage[accsupp]{axessibility}  

\makeatletter
\@namedef{ver@everyshi.sty}{}
\makeatother

\usepackage{bbm}
\newcommand{\compresslist}{%
\setlength{\itemsep}{1pt}%
\setlength{\parskip}{0pt}%
\setlength{\parsep}{0pt}%
}
\newcommand{\figref}[1]{Fig\onedot~\ref{#1}}
\newcommand{\equref}[1]{Eq\onedot~\eqref{#1}}
\newcommand{\secref}[1]{Sec\onedot~\ref{#1}}

\newcommand{\tabref}[1]{Tab\onedot~\ref{#1}}
\newcommand{\algref}[1]{Alg\onedot~\ref{#1}}

\usepackage{enumitem}

\newcommand{\Task}{Visual Planning for Assistance\xspace}

\newcommand{\T}{VPA\xspace}

\newcommand{\Prompt}{Goal Prompt\xspace}
\newcommand{\Vhist}{Visual History\xspace}

\newcommand{\M}{VLaMP\xspace}

 \usepackage[stable]{footmisc}

\usepackage[pagebackref=true,breaklinks=true,letterpaper=true,colorlinks,bookmarks=false]{hyperref}

\usepackage{packages/_math}
\usepackage{booktabs}
\usepackage{multirow}
\usepackage{mathtools}
\usepackage{subfigure}
\usepackage{placeins}
\usepackage[ruled,linesnumbered]{algorithm2e}
\iccvfinalcopy

\SetCommentSty{mycommfont}

\newcommand{\PTLM}{PTLM\xspace}
\def\iccvPaperID{1937} 

\ificcvfinal\fi

\begin{document}

\title{Pretrained Language Models as Visual Planners for Human Assistance}

\author{
Dhruvesh Patel$^{1,2}$
\quad Hamid Eghbalzadeh$^{1}$
\quad Nitin Kamra$^{1}$\\
\quad Michael Louis Iuzzolino$^{1}$
\quad Unnat Jain$^{1}$\thanks{equal mentoring\quad $^\dagger$corresponding author}
\quad Ruta Desai$^{1}$\footnotemark[1]\hspace{0.4em}\footnotemark[2]\\
\normalsize{$^{1}$Meta\quad $^{2}$UMass Amherst}\\
\normalsize{\url{https://github.com/facebookresearch/vlamp}}
}

\maketitle
\ificcvfinal\thispagestyle{empty}\fi

\begin{abstract}
In our pursuit of advancing multi-modal AI assistants capable of guiding users to achieve complex multi-step goals, we propose the task of  `\Task (\T)'.
Given a succinct natural language goal, e.g., ``make a shelf", and a video of the user's progress so far, the aim of \T is to devise a plan, \ie a sequence of actions such as ``sand shelf", ``paint shelf", etc. to realize the specified goal.
This requires assessing the user's progress from the (untrimmed) video, and relating it to the requirements of natural language goal, \ie which actions to select and in what order?
Consequently, this requires handling long video history and arbitrarily complex action dependencies.
To address these challenges, we decompose \T into video action segmentation and forecasting.
Importantly, we experiment by formulating the forecasting step as a multi-modal sequence modeling problem, allowing us to leverage the strength of pre-trained LMs (as the sequence model). This novel approach, which we call \textbf{V}isual \textbf{La}nguage \textbf{M}odel based \textbf{P}lanner (VLaMP), outperforms baselines across a suite of metrics that gauge the quality of the generated plans.
Furthermore, through comprehensive ablations, we also isolate the value of each component -- language pre-training, visual observations, and goal information.
We have open-sourced all the data, model checkpoints, and training code.
\end{abstract}

\section{Introduction}
\label{sec:intro}

\begin{figure}
    \centering
    \includegraphics[width=\linewidth]{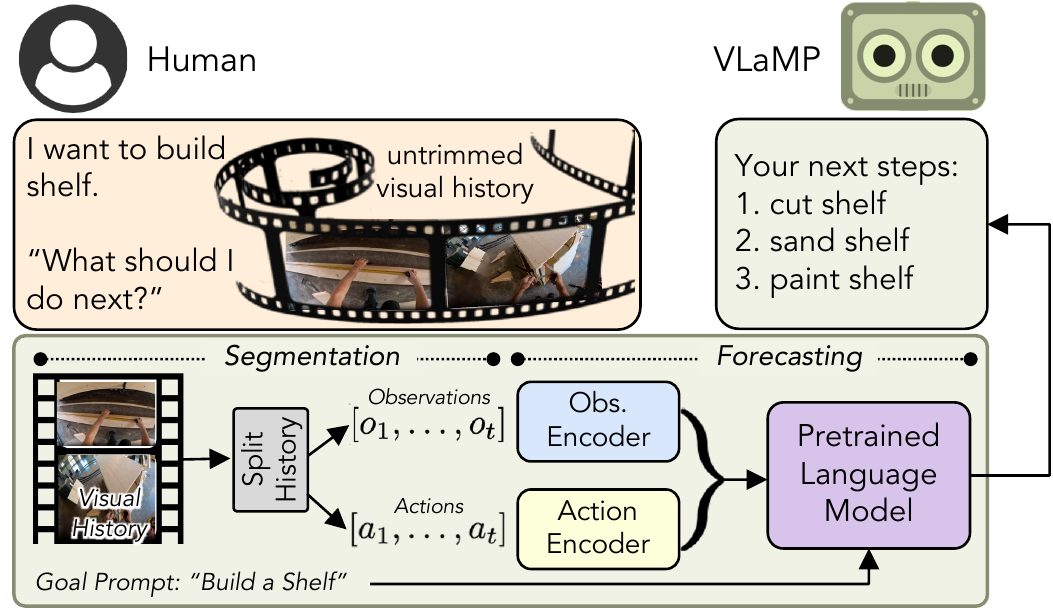}
    \caption[Caption for LOF]{\textbf{\Task overview (top) and general methodology (bottom).} Given a user-specified, natural language goal (``build a shelf'') and corresponding visual history of the user's progress till now, \T involves predicting a sequence of actions, to assist the user towards the goal (``cut'', ``sand'', ``paint''). Our approach is based on multi-modal sequence modeling where we reuse pre-trained video segmentation and language models.}
    \label{fig:teaser}
\end{figure}

Imagine assembling a new piece of furniture or following a new recipe for a dinner party. 
To achieve such a goal, you might follow 
a manual or a video tutorial, going back and forth as you perform the steps.
Instead of fumbling through a manual, imagine an 
assistive agent
capable of being invoked
through natural language, 
having the ability to understand human actions,
and providing actionable multi-step guidance for achieving your desired goal.
Such multi-modal assistive agents should be able to reason human activities from visual observations, contextualize them to the goal at hand, and plan future actions for providing guidance.

To quantify the progress and aid the development of such multi-modal neural models, we need a intuitive task, as illustrated in the example in~\figref{fig:teaser} (top). We call this \Task (\T) that we detail next.
Given a user-specified goal in natural language (``build shelf'')
and corresponding video observations of the user's progress towards this goal, the task objective should be to generate the ordered sequence of next actions  towards achieving the goal (``cut $\rightarrow$ sand $\rightarrow$ paint''). 
We base \T off instructional YouTube videos of such procedural activities from large, open-sourced datasets -- CrossTask~\cite{zhukov2019cross} and COIN~\cite{tang2019coin}. This is a natural choice as procedural human activities (cooking, assembly, repair, \etc.) are a perfect source of multi-step and complex sequence of actions, where humans routinely seek guidance. Realistic nature of \T makes it particularly challenging. Operating on untrimmed videos requires dealing with potentially many irrelevant background frames, chunking actions in the videos will also be imperative. Another challenge is the validity of the plan of actions \ie they must respect the constraints of the activity -- the shelf shouldn't be painted before sanding. 

The natural next question is -- what's a good way to tackle \T? Marrying research from video forecasting \& anticipation with embodied AI~\cite{duan2022survey,deitke2022retrospectives}, we cast \T as \textit{goal-conditioned} task planning.
We approach \T by utilizing video action segmentation and transformer-based neural sequence modeling -- the former allows us to deal with long video history and the latter can handle arbitrary sequential constraints \cite{yunAreTransformersUniversal2022, kratsiosUniversalApproximationConstraints2022}. The formulation (particularly, transformer-based neural sequence modeling) allows us to tap into the prowess of pre-trained language models (\PTLM{s}). These \PTLM{s} contain useful priors about action-action similarity, action-goal association, and action ordering~\cite{brohan2022can,huang2022language,pmlr-v162-paischer22a}, that our formulation can piggy-back off. Particularly, our model -- Visual Language Model Planner (\M) -- conditions the generated plan onto the visual history by using a transformer-based mapper network that projects embeddings corresponding to visual history into the input space of the LM. 
Using \T as the testbed, we show that \M outperforms a range of standardized baselines.
We undertake head-on ablations to quantify effect of each component of \M\ -- language pre-training of the LM, the visual history, and goal description. We believe the modularity of our approach can allow researchers to swap components with their own, to made rapid progress towards \T.

In summary, our main contributions are:
~(1)~a new task \T on human interaction videos capturing unique aspects of real-world vision-powered assistive agents;
~(2)~a general-purpose methodology for \T, which allows leveraging pre-trained multi-modal encoders and sequence models;
~(3)~an instantiation of this methodology (we call \M), where we reuse sequence priors from \PTLM and investigate its efficacy across two, well-studied datasets of procedural human activities.

\section{Related Work}\label{sec:related work}
\noindent\textbf{Forecasting in Videos.}
\T entails future action sequence prediction for a video and is closely related to anticipation in video understanding, including future localization~\cite{furnari2019would,lan2014hierarchical,vondrick2016anticipating},
future frame prediction~\cite{villegas2017decomposing,lotter2016deep,mendonca23swim}, next active object estimation~\cite{furnari2017next,bertasius2017first, grauman2022ego4d,Shan20,
bahl2023affordances,goyal2022human,
liu2022joint}, as well as short- and long-term action anticipation \cite{sener2019zero, damen2020rescaling, farha2020long, girdhar2021anticipative, grauman2022ego4d, sener2020temporal, sener2022transferring, nagarajan2020ego, loh2022long, abu2018will, mehrasavariational, mascaro2023intention, chen2022ego4d, wang2023ego}. While short-term forecasting approaches, such as \cite{sener2019zero}, are limited to predicting the single next action occurring only a few seconds into the future, our work focuses on predicting sequences of actions over longer time horizons (on the order of minutes) into the future. 
The long-term action (LTA) forecasting benchmarking was recently established on a small subset of Ego4D~\cite{grauman2022ego4d}, and although \T is similarly devised to predict a temporally ordered sequence of actions conditioned on video-based visual observations, our approach differs in that the LTA task does no goal-conditioning. Our task incorporates goal-conditioning on long term sequence prediction as we argue this will drive the development of critical system aspects, such as virtual assistants that afford human interaction via the natural language goals. Additionally, in contrast to LTA, we focus on a wide range of goal-oriented activities available within the CrossTask and COIN datasets. \tabref{tab:related_work} highlights the differences between \T and other forecasting tasks.

\begin{table}[]
\resizebox{\columnwidth}{!}{%
\begin{tabular}{@{}llll@{}}
\toprule
 &
  \begin{tabular}[c]{@{}l@{}}Action\\ Prediction\end{tabular} 
  & \begin{tabular}[c]{@{}l@{}}Goal\\ Modality\end{tabular}
  & \begin{tabular}[c]{@{}l@{}}Visual \\ Reasoning\end{tabular} \\ \midrule
Action anticipation~\cite{damen2018scaling} & Single            
& None  & Video-based \\
Action forecasting~\cite{grauman2022ego4d}  & Multiple, ordered          
& None  & Video-based \\
Procedural planning~\cite{chang2020procedure} & Multiple, ordered 
& Vision & Image-based \\
\textbf{\T (ours)}                 & Multiple, ordered 
& Natural Language & Video-based \\ \bottomrule
\end{tabular}%
}
\caption{\textbf{\T \vs prior tasks related to forecasting of real-world human activities.} We predict an ordered sequence of actions given video history with a focus on natural language goal-conditioning for human-assistive applications.
}
\label{tab:related_work}
\end{table}

\noindent \textbf{Procedural Planning Approaches.}
\T is similar to the task of procedure planning~\cite{chang2020procedure}, wherein given a starting and terminating visual observation, the aim is to predict the actions that would transform the state from starting to terminating.
However, we argue that due to the unavailability of the terminating visual state, the procedure planning task is not useful in a real-world assistance setting.
While several recent works introduce novel models for procedure planning, these models cannot be deployed for \T either because they rely heavily on the availability of visual goal~\cite{sun2022plate,zhao2022p3iv,bi2021procedure}, or assume access to true action history \cite{mao2023adtg}.

\noindent \textbf{Transformers for Decision Making.} \M autoregressively predicts future states and/or observations and actions and is similar in spirit to sequence models for decision making such as GATO~\cite{reed2022generalist}, Decision Transformer~\cite{chen2021decision}, and Trajectory Transformer~\cite{janner2021offline}. \M extends such models to work with egocentric observations that humans observe in the day-to-day activities. 

\noindent \textbf{Planning in Embodied AI using LMs.} 
Past research has leveraged \PTLM{s} for task planning in real world for embodied agents~\cite{huang2022language, singh2022progprompt, li2022pre, brohan2022can, liang2022code, huang2022inner, driess2023palme, dasgupta2023collaborating, jiang2022vima}. 
Many of these works focus on converting a high-level task or instruction into sequence of low-level steps and then ground them in the environment using either affordance functions~\cite{brohan2022can}, visual feedback~\cite{huang2022inner}, or multimodal prompts~\cite{jiang2022vima, driess2023palme}. 
Owing to their focus on robotic agents, these works focus on predominately pick-place tasks. Instead \T is focused on complex tasks, in which humans might require assistance in the form of recommended future actions. Consequently, \M requires grounding and reasoning of much more complex states and actions, which it accomplishes by finetuning a pretrained LM on multimodal sequences of (visual) observations and (text-based) actions.

\noindent \textbf{Multi-modal LMs.} 
Recent works have successfully used the transformer architecture for modeling multi-modal sequences that have visual tokens. 
For instance, ~\cite{lu2019vilbert,pashevich2021episodic,aghajanyan2022cm3,lu2022unified,huang2023language, driess2023palme} train large transformer based multi-modal sequence prediction models. 
While ~\cite{alayrac2022flamingo,tsimpoukelli2021multimodal,mokady2021clipcap} focus on adapting pre-trained LMs to work with visual tokens by aligning the representation spaces of the two modalities. 
\M's approach of modeling sequence of visual and textual representations using \PTLM{s} is similar in spirit to these, \ie \M also uses a mapper network to align video representations to LM's token space and jointly learns the mapper with LM finetuning for \T. 
However, in contrast to previous works, \M predicts the visual tokens autoregressively at inference time to enable forecasting of the state for planning. 
Consequently, \M's token prediction loss is multi-modal, unlike most multi-modal LMs that only use token prediction loss for text tokens.

\section{\Task}
\label{sec:background}
Here, we introduce the task of \Task(\T), towards enabling multi-step guidance to humans in their real-world activities. 
We instantiate \T for procedural activities, where humans routinely seek assistance.
In this section, we include the definition of \T, and describe the evaluation protocol. 
\subsection{Task Definition}
\label{subsec:task-def}
The following two intuitive inputs are given to any model performing \T. 

\noindent\textbf{\Prompt ($G$).} The natural language description (in short phrase) of the user's goal, emulating a typical user's request for assistance for a day-to-day task. Examples include, ``build a shelf'' and ``change a tire". 

\noindent\textbf{\Vhist ($V_t$).} 
An untrimmed video that provides context about the user's progress towards a goal from the start till time, say $t$. 
We assume that $V_t$ contains $k$ actions or steps $\{a_1, \dots, a_k\}$ pertaining to the goal. However, VPA doesn't have access to $\{a_1, \dots, a_k\}$ or $k$ and must work with $V_t$.

Given these two inputs, $V_t$ and $G$, the objective of \T is to generate a \textit{plan} $\mathcal{T}$. The plan is a
sequence of actions that should be executed (in the next steps) to assist the user in achieving the goal $G$. 
Concretely, the prediction is denoted by $\mathcal{T} = \left(a_{k+1},\dots,a_{k+l}\right)$, where $a_i$ are represented in natural language but come from a closed set $\mathcal{A}$. 
Here, $l\leq K$ denotes the number of future actions that should be predicted, out of the $K$ number of remaining actions required to accomplish the goal. 
For our shelf-building example, the correct $\mathcal{T} = \left(\mathtt{sand\:shelf},\;\mathtt{paint\:shelf},\;\mathtt{attach\:shelf}\right)$, capturing the remaining $3$ future actions (here, $l=N=3$).   
In day-to-day activities, we request assistance for goals in natural language. 
However, prior works \textit{procedural planning}~\cite{chang2020procedure,bi2021procedure} assume access to visual goal state.
This is not a realistic assumption for an AI agent assisting humans. 
Hence, we purposefully relax this assumption, making our formulation of the task significantly more practical.
Moreover, an agent with access to the visual modality,  should be able to improve its plan by filtering out relevant information from the raw video stream of the progress.
These are the two central assumptions around which we formulate the task of \T.

\subsection{Evaluation}
\label{subsec:evaluationofT}
\noindent\textbf{Open-Sourced Video Data.} 
We leverage existing datasets CrossTask~\cite{zhukov2019cross} and COIN~\cite{tang2019coin}, originally developed to enable video action understanding for \T, based on the following three requirements:

\noindent$\bullet\,$ \textit{Rich diversity of activities from multiple domains}: The data from different domains such as cooking, assembly etc.,  enables testing of \T models in a more generalized manner.

\noindent$\bullet\,$ \textit{Goal-oriented activities consisting of long sequences of actions}: Since the objective is to generate $l$ future actions, activities in these datasets that require diverse sequences of actions e.g., ``making a pancake",  instead of ``running" are more suitable for \T. 

\noindent$\bullet\,$ \textit{Action annotations from a fixed closed set of actions }: 
Action labels described using verb-noun from a finite set~\cite{damen2018scaling}, which are temporally aligned with the videos. 
This makes evaluating the accuracy of $\mathcal{T}$ prediction straightforward. 
Specifically, free-form, narration-style descriptions of actions that are available in recent video datasets aid in efficient multi-modal representation learning for video understanding~\cite{grauman2022ego4d,miech2019howto100m}. 
However, evaluating the quality of action \emph{sequences} towards goal achievement, where each action is described in free-form natural language, is non-trivial. We leave the instantiation of \T with free-form natural language actions as future work. 

Table~\ref{tab:data-stats} summarizes the features from CrossTask and COIN, aligned with the above requirements. 
\begin{table}[]
\centering
\resizebox{\columnwidth}{!}{%
\begin{tabular}{@{}llccccc@{}}
\toprule
Dataset &
  \begin{tabular}[c]{@{}l@{}}\# train \\ videos\end{tabular} &
  \begin{tabular}[c]{@{}c@{}}\# test \\ videos\end{tabular} &
  \begin{tabular}[c]{@{}c@{}}\# test \\ samples\end{tabular} &
  \begin{tabular}[c]{@{}c@{}}actions per\\ video\end{tabular} &
  \# {goals} &
  \# domains \\ \midrule
CrossTask~\cite{zhukov2019cross} &
  1756 &
  752 &
  4123 &
  7.6 ± 4.3 &
  18 &
  3 \\
COIN~\cite{tang2019coin} &
  9428 &
  1047 &
  2011 &
  3.9 ± 2.4 &
  180 &
  12 \\ \bottomrule
\end{tabular}%
}
\caption{\textbf{\T datasets.} We evaluate \T on two existing video datasets containing multiple goal-oriented procedural activities from varied domains. Such activities contain sequences of multiple actions making them ideal for \T.} 
\label{tab:data-stats}
\end{table}

\paragraph{Metrics.}
The planning performance of a \T model is measured by comparing the generated plan $\mathcal {\hat T}=(\hat a_{k+1}, \dots, \hat a_{k+l})$ to the ground truth plan $\mathcal{T}$ for $l$ actions in the future, given $V_t$ and $G$. Here $\hat a_{k+i}$ denotes the prediction for the $k+i$-th step given history till $k$-th step.
Consistent with community practices~\cite{chang2020procedure, bi2021procedure}, we use 
the following metrics, listed in decreasing order of strictness: \emph{success rate} (SR), \emph{mean accuracy} (mAcc), \emph{mean intersection over union} (mIOU). Success rate requires an exact match between all actions and their sequence between $\mathcal{\hat T}$ and $\mathcal{T}$. Mean accuracy is the
accuracy of the actions at each step. Unlike success rate, mean accuracy  does not require a 100\% match to ground truth. Instead it considers matching at each individual step. Lastly, mean intersection over union captures the cases where the model predicts
the steps
correctly, but fails to identify the correct order. Concretely,

\begin{align}
    \mathrm{mIOU}_l &= \frac{\left|\hat a_{\{k+1:k+l\}} \bigcap a_{\{k+1:k+l\}}\right|}{\left|\hat a_{\{k+1:k+l\}} \bigcup a_{\{k+1:k+l\}}\right|},\\
    \mathrm{mAcc}_l &= \frac{1}{l} \sum_{i=1}^{l} \mathbbm{1}[\hat a_{k+i,} = a_{k+i}],\\
    \mathrm{SR}_l &= \prod_{i=1}^{l}\mathbbm{1}[\hat a_{k+i} = a_{k+i}],
\end{align}
where $\mathbbm{1}[\cdot]$ is the identity function, which is $1$ when the condition in its input is true, and $0$, and $\hat a_{\{k+1:k+l\}}$ denotes the \emph{set} of $l$ future actions in $\mathcal{\hat T}$, i.e., a sequence but disregarding the order.
To complement the above metrics, we also measure the accuracy of predicting the next action \ie nAcc, as defined in \eqref{eq:nAcc}, where `n' stands for next. 
Note that all metrics are averaged over the test set details of which are included in~\secref{sec:experiments}. 
\begin{align}
    \mathrm{nAcc} = \mathbbm{1}[\hat a_{k+1} = a_{k+1}]
    \label{eq:nAcc}
\end{align}
\section{Visual LM Planner}\label{sec:model}
\T can be viewed as a sequential decision making problem, where the model (say, $\pi$) predicting the sequence of next actions is a policy conditioned on the visual history $V_t$, serving as (partially-observed) state, and goal prompt $G$.
Inspired by the offline learning formulation closest to `learning from offline demonstrations'~\cite{hester2018deep,vecerik2017leveraging,kang2018policy,Weihs2020Bridging}, in \M, we formulate \T as a goal-conditioned, multi-modal sequence prediction problem.
\footnote{Future works may explore alternate policy optimizations based on reward shaping, inverse RL, or by employing additional online interactions through photorealistic simulation~\cite{kolve2017ai2,habitat19iccv,szot2021habitat,srivastava2022behavior,AllenAct,CSync,TBP,chen2020soundspaces,wani2020multion}.}
This formulation allows us to leverage high-capacity sequence models like Transformers~\cite{vaswani2017attention}, which have been successfully applied to sequential decision making ~\cite{janner2021offline,chen2021decision}.
Furthermore, inspired by the recent success of pre-trained transformer language models (PTLMs) 
on such tasks,
~\cite{sakaguchi2021proscript, huang2022language,chen2021decision}, we propose to leverage PTLMs as the sequence model in our formulation.
In the remainder of this section we describe our approach for \T, called  \textbf{V}isual \textbf{La}nguage \textbf{M}odel based \textbf{P}lanner (VLaMP), consisting of a segmentation module and PTLM based sequence prediction module. 

\subsection{Planning with Segmentation and Forecasting}

We define $\pi$ as a goal-conditioned, multi-modal sequence prediction problem:
\begin{align}
    \pi = P (a_{k+1}, a_{k+2}, \dots \mid V_t,G).
    \label{eq:vpa}
\end{align}
where $\pi$ models the probability of goal-relevant and valid future actions sequences conditioned on $V_t$ and $G$. 

Modeling the multi-modal sequence in Eq.~\ref{eq:vpa} is computationally expensive and difficult to scale because of the high-dimensional state space of raw untrimmed video $V_t$.\footnote{A typical untrimmed video of 100p video of 5 minutes at 8fps will have $2^7$ raw pixel values.}
Also, there is limited data to learn the distribution over valid action sequences for goals in real-world applications. 
Tackling these challenges, we argue that the latent space of factors influencing the future plan can indeed be expressed in lesser dimensions.
Particularly, to ensure scalability for handling raw videos and sample efficiency in learning valid action sequence distributions, we \textit{decompose} our policy $\pi$ into two modules.
The first module is \emph{video segmentation}, which converts the untrimmed video history $V_t$ into a sequence of video segments \ie a segment history $S_k=(s_1, \dots, s_k)$, where each segment corresponds to an action $a_i$ that occurred in the video.
The second module 
enables
\emph{forecasting} \ie 
it transforms
the output of the segmentation module and generates the plan (see Fig.~\ref{fig:teaser} (bottom)).
While a probabilistic formulation of this decomposition can be expressed as: 
\begin{align}
    \pi =   \sum_{S_k}\underbrace{P(a_{k+1}, a_{k+2}, \dots \mid S_k, G)}_{\textrm{Forecasting}}\underbrace{P(S_k\mid  V_{t})}_{\textrm{Segmentation}},
    \label{eq:framework}
\end{align}
where the segment history $S_k$ is a latent variable, the summation over all possible segment histories is intractable.
We, however, use Eq. \ref{eq:framework} as the guiding expression to formulate the input and output of both modules.  
Next, we include the technical details of both of these modules. 

\begin{figure}
    \centering
    \includegraphics[width=\columnwidth]{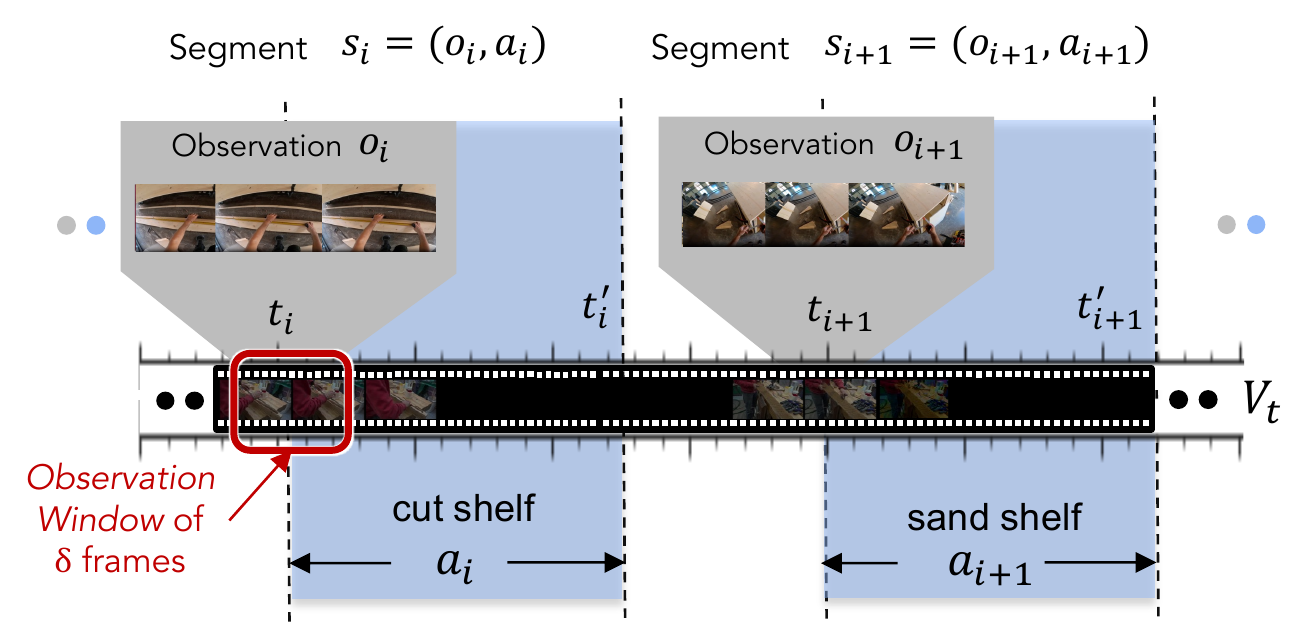}
    \caption{\textbf{\M~--~Segmentation Module.} The untrimmed visual history $V_t$ is converted into segments, each consisting of observation $o_i$ and the action $a_i$. The observation $o_i$ is the collection of video frames of $\delta$ seconds around the start time stamp $t_i$ of the corresponding action $a_i$. Two such segments are shown here.
    }
    \label{fig:segmentation}
\end{figure}

\subsection{Segmentation Module}\label{sec:segmentation module}
This module splits the untrimmed video history $V_t$ into segment history $S_k$ of multiple segments, each segment corresponding to an action. The segmentation is done using a video-action segmentation model in three steps: pre-processing, classification, and consolidation.
In the pre-processing step, the raw video frames from $V_t$ are bundled into fixed-length window clips $c_i$, each of length 1 second, to obtain $V_t = (c_1, \dots, c_t)$.
In the classification step, a video-action segmentation model is used to output the most probable action for each clip $c_i$, which can be denoted as $\tilde A_t = (\tilde{a}_1, \tilde a_2,\dots, \tilde a_{t})$.
Finally, in the consolidation steps,  we convert $\tilde A_t$ into a form that can be used by the forecasting module; this form consists of two sequences: \textbf{action history} $A_k$ and \textbf{observation history} $O_k$. 
To this end, same actions in consecutive seconds in $\tilde A_t$ are consolidated to form the action history $A_k=(a_1, \dots, a_k)$. As illustrated in \figref{fig:segmentation}, assuming $t_i$ denotes the starting timestamp for $a_i$, we also extract video frames from $t_i-\delta/2$ to $t_i+\delta/2$ to obtain a 
\textit{observation window} $o_i$ corresponding to $a_i$, and consequently the full observation history $O_k = (o_1, \dots, o_k)$.
The resultant segment history is termed $S_k=((o_1, a_1), \dots, (o_k, a_k))$, summarized in \figref{fig:segmentation}.

%
%
%
%
%
%
%
%
%
%
%
 \subsection{Forecasting Module}\label{sec:forecasting model}
\begin{figure*}
    \centering
    \includegraphics[width=0.9\linewidth]{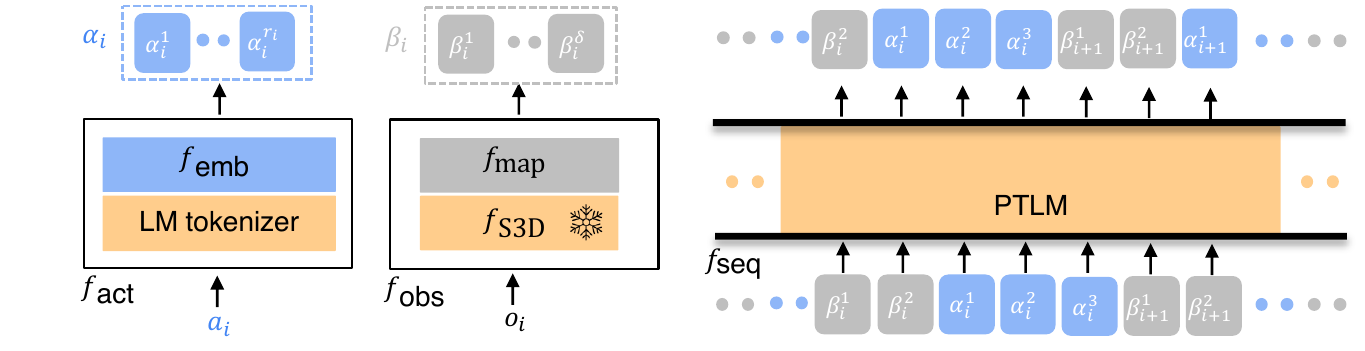}
    \caption{\textbf{\M~--~Forecasting Module.}~As shown in left and middle, actions and observations obtained from the segmentation module are encoded using appropriate modality encoders. The observation encoder leverages pretrained video encoder for observations, while also learning a mapper that aligns the representations from observations with actions. As shown on the  right, VLaMP uses a joint sequence model on top of interleaved action (blue) and observation (gray) representations to forecast autoregressively the next representation.}
    \label{fig:forecasting model}
\end{figure*}

The usefulness of $\pi$'s decomposition expressed in \equref{eq:framework} becomes apparent now. 
Modeling the  segmentation module's output as segment history $S_k$, where
each segment consisting of action and observations, allows writing the output of the forecasting module in an autoregressive manner:
\begin{align}
    &P(a_{k+1}, a_{k+2}, \dots \mid o_1, a_1, \dots, o_k, a_k, G)\nonumber\\
    &= \prod_{i>0}\sum_{o_{k+i}}P(o_{k+i}, a_{k+i}\mid o_1, a_1, \dots, o_k, a_k, G).
    \label{eq:autoreg}
\end{align}
Illustrated in \figref{fig:forecasting model},
this autoregressive expression allows the possibility of using any sequence-to-sequence neural network as the forecasting module in combination with pretrained text and video encoders for representing action and observation history $A_k$ and $O_k$.
This general-purpose framework allows the use of any neural sequence model -- LSTM \cite{hochreiter1997long}, GRU~\cite{cho2014properties}, Transformers \cite{vaswani2017attention}, \etc.
We choose to instantiate the forecasting module for $\pi$ using a pretrained transformer-based LM.
Next we present the details of the encoders for the two modalities and the LM based sequence model.

\noindent \textbf{Action encoder ($\AEnc$)}\label{sec:action encoder} 
Each action $a_i$ in $A_k$ is encoded by $\AEnc$ and the output is denoted by $\mathbf{\alpha}_i$. 
Concretely, token embeddings are expressed as:
\begin{align}
    (\alpha_1, \dots, \alpha_k) &= \left(\AEnc(a_1), \dots, \AEnc(a_k)\right), \text{where}\nonumber\\
    \mathbf{\alpha}_i &= (\alpha_{i}^1, \dots, \alpha_i^{r_{i}}) \in \mathbb R^{r_i \times d}
\end{align}
As we illustrate in \figref{fig:forecasting model}~(left),
each action $a_i$ is tokenized into $r_i$ tokens using appropriate tokenizer for the LM, the tokens are indexed using the vocabulary of the LM, and are represented using an embedding lookup from the token embeddings of the LM to produce $\mathbf{\alpha}_i$. Here, $r_i$ is the number of tokens and $d$ is the dimensionality of token embeddings.

\noindent\textbf{Observation encoder 
($\VEnc$).
}
Visual cues play an integral role in knowing what actions lie ahead, towards achieving the goal prompt $G$. To this end, the visual observations $O_k$ are encoded and play a critical role in the planner. 
Recall, the visual observation history $O_k$ comprises of $o_i$ corresponding to action $a_i$, each of $\delta$ frames.
As illustrated in \figref{fig:forecasting model}~(middle), 
we transform each $o_i$ employing the widely-adopted S3D backbone~\cite{s3d} $f_{\textrm{S3D}^*}$ ($\ast$ denotes backbone is frozen).
We must project visual encodings to a shared latent space of action (language) embeddings described before ($\alpha_i$). To this end, we map S3D features via a trainable transformer mapper $f_{\textrm{map}}$. Concretely,
\begin{align}
    (\beta_1, \dots, \beta_k) &= \left(\VEnc(o_1), \dots, \VEnc(o_k)\right), \text{where}\nonumber\\
    \mathbf{\beta}_i = (\beta_{i}^1, \dots, \beta_i^{\delta}) &\in \mathbb R^{\delta \times d}\text{ and }\VEnc=f_{\textrm{S3D}^*}\circ f_{\textrm{map}}
\end{align}

\begin{figure}
    \centering
    \includegraphics[width=0.8\columnwidth]{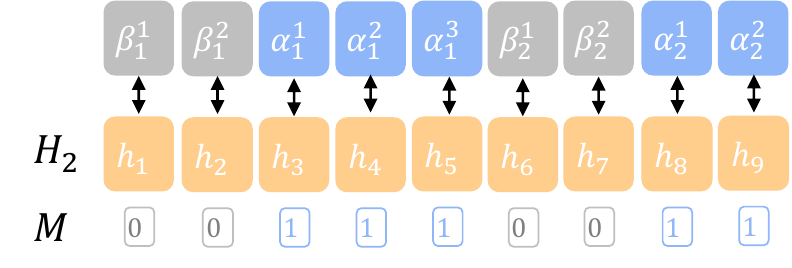}
    \caption{\textbf{Tokenized Sequence with Masks.} The encoded sequence of representations for $k=2$ segments, denoted alternatively using modality agnostic notation of $H_2$ and mask $M$ for next token prediction training.}
    \label{fig:H}
\end{figure}

Overall, as we show in \figref{fig:forecasting model} (right), the resultant encoded sequence of representation for $S_k$ is thus
$$f_{\textrm{enc}}(S_k) = (\beta_1, \alpha_1, \dots, \beta_k, \alpha_k)=H_k,$$
\noindent\textbf{Sequence model ($f_{\textrm{seq}}$).}
Given the above encoding for the segment history $S_k$, the role of the sequence model is to predict a representation of the next token, that would in return enable VLaMP plan generation capabilities for \T. 
Importantly, in the process of generating sequence of future actions autoregressively, we would also need to generate the  representations of `future observations'.
Therefore, as we shown in \figref{fig:forecasting model}, our sequence model, which consists of the transformer layers of a \PTLM, also produces representations for vision (in addition to the necessary action tokens).
Before proceeding further, we pause and introduce  additional notation for the sequence model, which will make the subsequent explanation for training and inference easier to follow.
As shown in Figure \ref{fig:H}, we alternatively denote the sequence of representations $(\beta_1, \alpha_1, \dots, \beta_k, \alpha_k)$ by $H_k=(h_1, \dots, h_{n})$, with $n=k\delta+\sum_{i=1}^k r_i$. 
A  binary mask $M=(m_1, \dots, m_n)$, where $m_i$ is $1$ if the corresponding representation is for an action and $0$ otherwise,  can help obtain necessary action or visual observations.
With this notation, given first $j$ representations denoted as $h_{1:j}$, one step of the sequence model produces the representation for $j+1$, i.e., $f_{\textrm{seq}}(h_{1:j}) = \hat h_{j+1}$.

\subsection{Training}\label{sec:training}
The joint training of the segmentation and forecasting modules following Eq. \eqref{eq:framework} is intractable.
\footnote{Despite tuning attempts, we found joint optimization to be intractable and inefficient on resources.} 
But, by exploiting the availability of unpaired training data, we approximate Eq. \eqref{eq:framework} by feeding in the output of the segmentation module to the forecasting module and training them separately, each on their respective labeled data. 
The video-action segmentation model is trained utilizing the VideoCLIP setup~\cite{xu2021videoclip}, where in the segmentation model performs classification to predict the action for each second of the video.
The forecasting model is trained by adopting the next representation prediction objective.
Unlike vanilla LM pretraining, however, we also need to train for predicting visual representations in addition to text (action).
Therefore, we use two different losses $ L_{\textrm{act}}$ and $ L_{\textrm{obs}}$ for text and visual representations respectively.
Specifically, 
$L_{\textrm{act}}$ is the conventional cross-entropy loss over the LM's vocabulary $V_{\textrm{LM}}$ for the action representations while  $L_{\textrm{obs}}$ is the mean-squared error between the predicted and the ground truth observation representations.
The total loss is the sum of both the loss terms as shown in \equref{eq:obs/act loss}.

\begin{align}
    L = -\sum_{j=1}^{n} & m_j\, L_{\textrm{act}}(\hat h_j) + (1- m_j)\,  L_{\textrm{obs}}(\hat h_j), \text{where}\nonumber\\
     L_{\textrm{act}} &=  h_j\cdot\hat h_j - \log \sum_{p=1}^{|V_{\textrm{LM}}|} \exp\left( h_p\cdot\hat h_j\right)~;\\
     L_{\textrm{obs}} &=  \frac{\Vert h_j - \hat h_j \Vert^2_2}{d}
    \label{eq:obs/act loss}
\end{align}

In order to have a stable training, we use ground truth action history to construct $S_k$ (and subsequently $H_k$) instead of the output of the segmentation module. 
Appendix \ref{app:training/opt/inference} provides further details on loss and optimizers for training.

\noindent\textbf{Inference.}
We next detail the inference procedure for \M. Recall that we use $1:n$ to denote a sequence of $n$ representations (i.e, $h_{1:n} = (h_1, \dots, h_n)$).
Additionally, 
we denote the concatenation operator over two representation sequences by ``$\diamond$''.
With this notation at hand, we define the $score$ of an action $a\in \mathcal A$ for following history $h_{1:n}$ as 
\begin{align}
    \phi(h_{1:n} \diamond f_{\textrm{act}}(a)) = \sum_{j=1}^{r_a} a^j \cdot f_{\textrm{seq}}(h_{1:n} \diamond a^{1:j}),
\end{align}
where $\cdot$ is the vector dot product, and $f_{\textrm{act}}(a)=\alpha^{1:r_a}=(\alpha^1, \dots, \alpha^{r_a})$ is the sequence of encoded representations for action $a$.
In other words, this \emph{score} is 
the sum of unnormalized log-probability under the sequence model using the standard softmax distribution. 
We use this scoring function with to perform beam search (detailed inference algorithm: Algo.~\ref{alg:inference-ext} in App.~\ref{app:training/opt/inference}).
\section{Experiments}
\label{sec:experiments}
We instantiate \M's segmentation module utilizing VideoCLIP \cite{xu2021videoclip} that has been fine-tuned on COIN and CrossTask. Similarly, we instantiate \M's sequence model $f_{\textrm{seq}}$ (in the forecasting module) by GPT2 \cite{gpt2}. We utilize the open-sourced model weights of GPT2 (from HuggingFace~\cite{wolf2020transformers}).
For inference, we use Algo.~\ref{alg:inference-ext} presented in Appendix \ref{app:training/opt/inference}, with beam size $B=10$ for CrossTask and a $B=3$ for COIN.

\subsection{Data and Baselines}\label{sec:data details}

\noindent\textbf{Data.} For a video $V$ with goal $G$, both CrossTask~\cite{zhukov2019cross} and COIN~\cite{tang2019coin} provide annotations of the form $\{a_k, (t_k, t_k')\}_{k=1}^K$, where $a_k$
are the actions in the video, and $t_k$ (resp. $t_k')$ are the start (resp. end) timestamps for $a_k$.~\footnote{Despite the presence of prepositions, majority of actions in COIN and CrossTask can be described using a verb-noun pair. While similar verbs exist, they are never paired with the same nouns. Therefore, there are no repeated actions. Example: while verb `lift' w/ noun `barbell' and verb `pickup' w/ `button' exists in the action library, the complements (`lift button' \& `pick up barbell') do not. }
Given an annotated video consisting of $K$ steps, we generate $K-l$ examples, each with input $x_k=(G, V_{t_k})$ and output $y_k=(a_{k+1}, \dots, a_{k+l})$, for $k=1, \dots K-l$ (leaving at least $l$ steps to predict in each example).
\footnote{Since we evaluate for three and four next steps on our datasets, we use the maximum required length and set $l=4$. }
Therefore from $M$ videos, we generated $N=\sum_{m=1}^M(K_m-l)$ examples, where $K_m$ is the number of steps in the $m$-th video, forming a dataset $\mathcal D=\{x^{(j)}, y^{(j)}\}_{j=1}^N$ suitable for \T (total number of samples in shown in Table \ref{tab:data-stats}).

\noindent\textbf{Baselines.}
As a first step towards benchmarking, we utilize two heuristic baselines -- a \emph{random} baseline and \emph{most probable action} baseline. Additionally, we also adopt a variety of strong goal-conditioned models.
The procedure planning task is most relevant task from the literature to \T, therefore, we adapt (details in App.~\ref{app:ddn_details})
the widely used DDN model introduced by Chang~\etal~\cite{chang2020procedure}, which is an established  model in many procedural planning benchmarks
in prior works~\cite{bi2021procedure,zhao2022p3iv,sun2022plate}. 
Building bridges to the research community working on Ego4D's Long Term action Anticipation benchmark (LTA)~\cite{grauman2022ego4d}, we also evaluate their best performing baseline. 
Finally, we include a prompt-based GPT-3 planner.
Zooming out, these baselines span several paradigms: \emph{prompt-based}, and \emph{random} baselines are learning-free, the \emph{most-probable} and \emph{random} baselines are non-neural, while \emph{DDN} and \emph{LTA} baselines are learned and neural.
Succinct descriptions of baselines are included next (details are deferred to Appendix~\ref{app:baselines}):

\noindent$\bullet$ \emph{Random}: Predicts the plan by picking all $l$ actions uniformly randomly from the set of all actions $\mathcal A$.\\
\noindent$\bullet$ \emph{Random w/ goal}: A stronger baseline; for each goal $G$, we allow privilege access to a set of applicable actions to that goal $\mathcal A_G \subseteq \mathcal A$, and predicts the plan by randomly picking actions from the restricted set.\\
\noindent$\bullet$ \emph{Most probable action}: 
Given the previous action $a_j$, picks the most probable next action $a_j$ according conditional probability $Pr(a_i|a_j)$ obtained using frequency count from the training. \\
\noindent$\bullet$ \emph{Most probable action w/ goal}: Akin to random w/ goal baseline, we also evaluate a goal-conditioned most probable action baseline, that uses a goal-specific set of actions $\mathcal A_G\subset \mathcal A$ during sampling. 
Since the most probable baselines, provide a probability distribution over the actions, we also employ beam search (for fairness, with the same beam size same as \M), and pick the highest scoring plan.\\
$\bullet$ \textit{Dual Dynamics Network (DDN)}~\cite{chang2020procedure}: Accounting for the difference in task definition, \ie lack of visual goal, a direct application of DDN inference algorithm was not possible. 
So we keep DDN's network structure but use Algo.\,\ref{alg:inference-ext} for inference on \T.\\
$\bullet$ \textit{Ego4D Long Term Action Anticipation~(LTA)~\cite{grauman2022ego4d}}~: 
The best performing baseline for LTA uses a SlowFast visual encoder followed by a transformer aggregator. We adapt this baseline using the S3D encoder for a fair apples-to-apples comparison with \M. 
\\
$\bullet$ \textit{Prompt-based GPT-3 Planner}: 
Huang~\etal~\cite{huang2022language} utilize a LLM as zero-shot planners, we also experiment with prompting a frozen pretrained large language model \ie GPT-3 for \T (additional details in App.~\ref{app:additional baseline GPT3 planner}). 
\\
%
%

%
%
%

%
\subsection{Quantitative Results} 
In the following, we include quantitative findings of benchmarking methods on two video data sources (\tabref{tab:main-results-vlamp}) and head-on ablations (\tabref{tab:ablations-short}).

\noindent\textbf{Improved performance across video datasets.} 
As we show in \tabref{tab:main-results-vlamp}, \M significantly outperforms the baselines for both CrossTask and COIN. DDN that is customized for procedural tasks performs significantly better than heuristic baselines leading to a mAcc boost from $12.7\rightarrow24.1\%$ (row 2 and 5, $l=4$, \tabref{tab:main-results-vlamp}).
With our novel decomposition and pretraining objective, \M outperforms DDN with a further bump up from $24.1\rightarrow31.7\%$ (row 5 and 7).

\noindent\textbf{Steady gains in short \& long horizon predictions.} As length of prediction $l$ increases ($l=1$ to $l=4$), the performance naturally decreases, across all baselines and tasks. With this in mind and zooming in on COIN results, we observe large gains of \M over DDN (next best method). Particularly, a relative improvement of $54\%$ ($29.3\rightarrow45.2$) in mAcc for $l=1$ (row 7 and 8, \tabref{tab:main-results-vlamp}) and  $68\%$ ($21.0\rightarrow35.2$) for $l=4$ is demonstrated by \M over DDN.

\begin{table}[]
\centering
\setlength{\tabcolsep}{3pt}
\resizebox{\columnwidth}{!}{%
\begin{tabular}{@{}lrrrrrrr@{}}
\toprule
  \multirow{2}{*}{\textbf{Method}} &
  $l=1$ &
  \multicolumn{3}{c}{$l=3$} &
  \multicolumn{3}{c}{$l=4$} \\ \cmidrule(l){2-8} 
   &
  \{n/m\}Acc &
  SR &
  mAcc &
  mIOU &
  SR &
  mAcc &
  mIOU \\ \midrule
\multicolumn{8}{c}{CrossTask Video Dataset~\cite{zhukov2019cross}}\\
\midrule
  {Random} &
  0.9   &
  0.0   &
  0.9   &
  1.5   &
  0.0   &
  0.9   &
  1.9  \\
  {Random w/ goal} &
  13.2  &
  0.3   &
  13.4  &
  23.6  &
  0.0   &
  12.7  &
  27.8  \\
 Most probable &
  10.4  &
  1.7  &
  6.1  &
  9.9  &
  1.3  &
  5.5  &
  13.9  \\
  Most probable w/ goal &
  12.4  &
  2.4  &
  8.9  &
  15.5  &
  1.5  &
  7.9  &
  20.5  \\
  DDN~\cite{chang2020procedure} &
  33.4  &
  6.8   &
  25.8  &
  35.2  &
  3.6   &
  24.1  &
  37.0  \\
  LTA~\cite{grauman2022ego4d} &
  26.9   &
  2.4   &
  24.0   &
  35.2  &
  1.2   &
  21.7   &
  36.8 \\
  VLaMP (ours) &
  \textbf{50.6} &
  \textbf{10.3} &
  \textbf{35.3} &
  \textbf{44.0} &
  \textbf{4.4} &
  \textbf{31.7} &
  \textbf{43.4} \\ \midrule
\multicolumn{8}{c}{COIN Video Dataset~\cite{tang2019coin}}\\
\midrule
  {Random} &
  {0.1}   &
  {0.0}   &
  {0.1}   &
  {0.2}   &
  {0.0}   &
  {0.1}   &
  {0.2} \\
  {Random w/ goal} &
  {24.5}   &
  {1.7}   &
  {21.4}   &
  {42.7}  &
  {0.3}  &
  {20.1}  &
  {47.7}   \\
  Most probable &
  0.7  &
  1.6  &
  4.3  &
  6.8  &
  1.6  &
  8.2  &
  15.3  \\
  Most probable w/ goal &
  23.9  &
  10.9  &
  18.0  &
  24.9  &
  9.1  &
  16.3  &
  32.2  \\
  DDN~\cite{chang2020procedure} &
  29.3   &
  10.1   &
  22.3   &
  32.2   &
  7.0   &
  21.0   &
  37.3   \\
   GPT-3 prompt-based ~\cite{huang2022language} &
  19.3   &
  1.7   &
  11.1   &
  19.5   &
  0.0   &
  10.5   &
  21.0   \\
  VLaMP (ours) &
  \textbf{45.2} &
  \textbf{18.3} &
  \textbf{39.2} &
  \textbf{56.6} &
  \textbf{9.0}  &
  \textbf{35.2} &
  \textbf{54.2} \\ \bottomrule
\end{tabular}%
}
\caption{\textbf{Performance on different datasets and horizons.} The mean of various metrics (Sec.~\ref{subsec:task-def}) obtained using 5 runs with different random seeds (std. errors are provide in Appendix \ref{app:more_quant}).
Note that the action and observation history are the output of the separately finetuned video-action segmentation model and hence are noisy compared to the ground truth history.}
\vspace{-3mm}
\label{tab:main-results-vlamp}
\end{table}
\noindent \textbf{Goal-conditioning is crucial.}~
A key difference between the task formulations of LTA and \T is goal conditioning (see Sec.~\ref{sec:related work}). 
Comparing rows for LTA model and \M -- there is almost 2x gap (27\% \textit{vs.}\ 51\%, l=1, Tab.~\ref{tab:main-results-vlamp}). This underscores the importance of goal-conditioning.
To second this inference, in \tabref{tab:ablations-short} (rows 1 and 2), we precisely ablate the effect of providing the goal as a \textit{textual description}
for the basic (last-observation-only) model. The only difference is goal prompt $G$, which increases mAcc performance from $44.5\rightarrow53.1\%$ for $l=1$ and $28.3\rightarrow34.7\%$ for $l=3$.

\noindent\textbf{Goal-conditioned random {and most probable} baseline come close to DDN on easy metrics.}  
For fair comparisons, we include heuristic baselines. We observe the performance of \emph{random w/ goal} and \emph{most probable w/ goal} comes close to DDN, when evaluated using lenient metrics like mAcc and mIOU (see COIN results in \tabref{tab:main-results-vlamp}).
Note, that these two baselines enjoy the privileged access to (a much smaller) `relevant actions set' for a given goal. On an average, the relevant or feasible action set is smaller for COIN than CrossTask videos. This apriori access to feasible actions make \emph{random w/ goal} and \emph{most probable w/ goal} competitive (albeit, slightly unfair) baselines.

\noindent\textbf{Prompt-based LLM planner is not competitive.}
{
We evaluate the prompt-based GPT-3 planner on COIN and find that the prompt based model performs significantly worse than \M. This highlights that \T is not easy to solve using just a PTLM and prompting. 
}

\subsection{Ablations and Error Analysis}\label{sec:error analysis}
We undertake head-on ablations to quantify effect of each component of \M\ -- action history, the visual history, and language pre-training of the LM.
To remove confounding factors, in \tabref{tab:ablations-short} we use the ground truth output for the segmentation module. Following this, we include detailed error analysis.
\begin{table}[]
\centering
\resizebox{\columnwidth}{!}{%
\begin{tabular}{@{}lccccccc@{}}
\toprule
                & $G$       & $A_k$     & $O_k$     & \multicolumn{3}{c}{$l=3$}                                                      & $l=1$                    \\ \midrule
                &           &           &           & SR                       & mAcc                     & mIOU                     & nAcc                     \\ \midrule
{\scriptsize 1} &
  \ding{55} &
  \ding{55} &
  $o_k$ &
  6.8 {\scriptsize ± 0.3} &
  \multicolumn{1}{l}{28.3 {\scriptsize ± 1.9}} &
  \multicolumn{1}{l}{34.8 {\scriptsize ± 2.0}} &
  44.5 {\scriptsize ± 3.8} \\
{\scriptsize 2} & \ding{51} & \ding{55} & $o_k$ & 8.9 {\scriptsize ± 0.2}  & 34.7 {\scriptsize ± 0.7} & 41.6 {\scriptsize ± 0.8} & 53.1 {\scriptsize ± 2.0} \\
{\scriptsize 3} & \ding{51} & \ding{51} & \ding{55} & 14.9 {\scriptsize ± 0.3} & 37.8 {\scriptsize ± 0.4} & 50.8 {\scriptsize ± 0.6} & 48.0 {\scriptsize ± 0.2} \\
{\scriptsize 4} & \ding{51} & \ding{51} & $O_k$ & 15.2 {\scriptsize ± 0.3} & 43.5 {\scriptsize ± 0.8} & 51.4 {\scriptsize ± 0.9} & 64.8 {\scriptsize ± 0.9} \\ \midrule
\textbf{R}      & \ding{51} & \ding{51} & $O_k$ & 10.7 {\scriptsize ± 0.2} & 36.5 {\scriptsize ± 0.7} & 41.7 {\scriptsize ± 0.6} & 61.4 {\scriptsize ± 1.8} \\ \bottomrule
\end{tabular}%
}
\caption{\textbf{Role of different inputs and LM pre-training in \M.}  $G$, $A_k$, $O_k$ denote the three inputs to \M: goal, action history and observation history, respectively. Shorthand of $o_k$ means only the most recent observation of the full history ($O_k$) is used. Mean $\pm$ std. error over 5 random seeds on CrossTask are reported.
The row \textbf{R} corresponds to the forecasting module of \M trained with random initialization (same model architecture) as opposed to initialization using the weights of a pre-trained LM.}
\label{tab:ablations-short}
\end{table}

\noindent\textbf{Action and observation history improve complementary planning metrics.}
As seen in all the rows with action history \tabref{tab:ablations-short} (\ie~$A_k$), the provision of action history increases difficult metrics such as SR. In comparing rows 2 and 3, we see that SR increases relatively by 67\% ($8.9\rightarrow14.9$), simply by using action history even without access to any past observation. However, the lack of observation history affects nAcc, which drops relatively by 10\% ($53.1\rightarrow48.0$) when observation history is swapped by action history between rows 2 and 3. This implies that observation history is important to predict the person's state in the task and what they might do next. 

\noindent\textbf{Priors from the pre-trained LM improve performance.}
In Tab.~\ref{tab:ablations-short} row \textbf{R}, we report the performance of \M when its forecasting module is trained with random weight initialization, \ie, the transformer of the same architecture trained from scratch instead of LM pre-training. We find the performance in row $\textbf{R}$ is thus much lower than that of \M with pre-trained LM shown in row 4. This underscores the crucial importance of LM pre-training.

\noindent\textbf{Segmentation errors are detrimental}.
While the accuracy of video-action segmentation module (particularly, VideoCLIP model) is quite good -- 80.2\% for CrossTask (68.7\% for COIN), we believe this should be a major focus for future research. Why?
The effect of this segmentation error in \M's  performance is quantified by comparing \M row in Tab.~\ref{tab:main-results-vlamp} (CrossTask) with corresponding performance assuming perfect segmentation (Tab.~\ref{tab:ablations-short}, row 4). A relative 50\% gap in success and big gaps in all other metrics.
\begin{figure}[h]
    \centering
    \includegraphics[width=0.85\columnwidth]{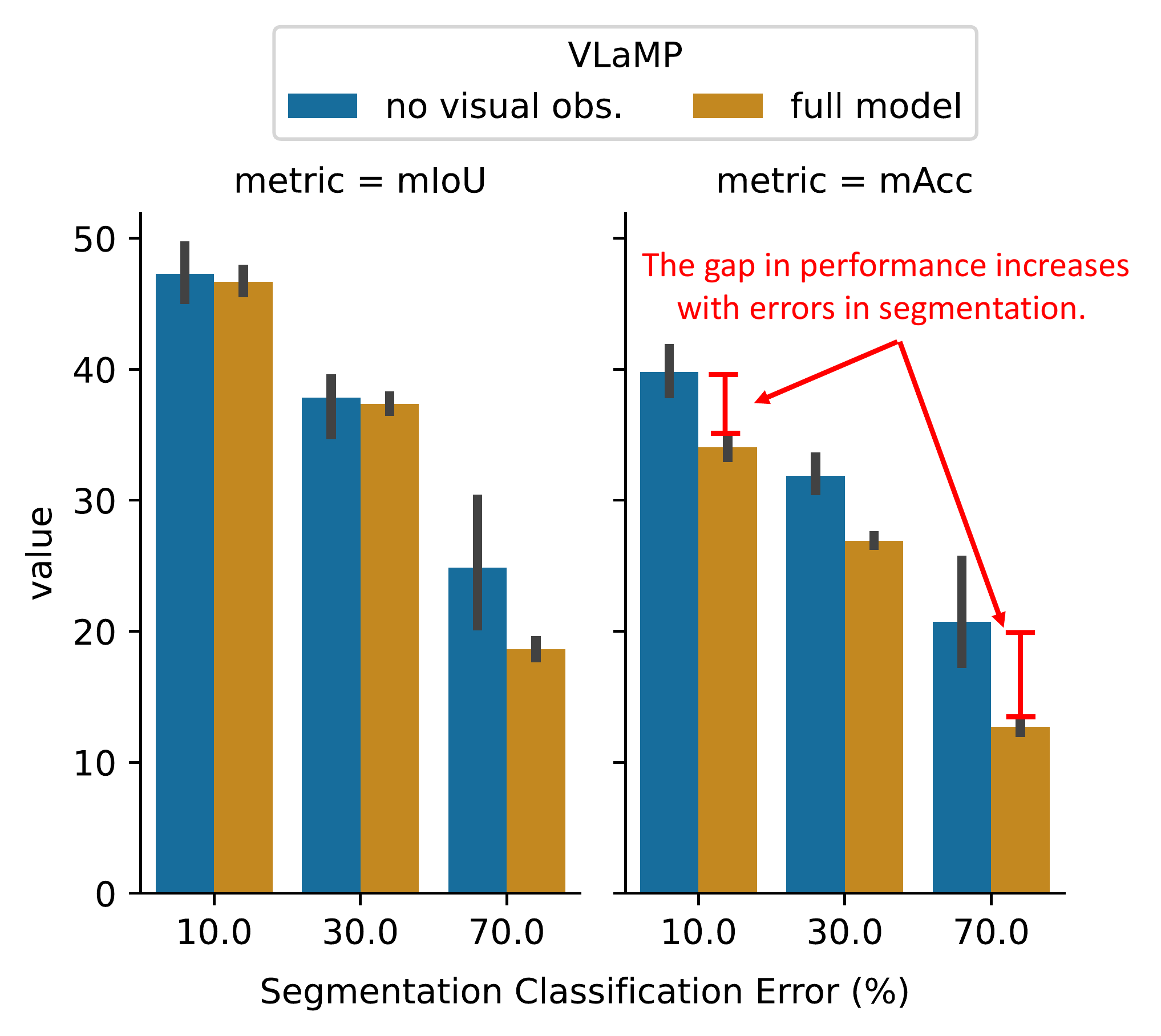}
    \vspace{-3mm}
    \caption{\textbf{Effect of segmentation errors.} Performance of VLaMP on CrossTask with classification errors (\%) in the video-action segmentation. As the errors increase, the performance gap between the full model and the one without access to observations increases.}
    \label{fig:random noise}
\end{figure}

\noindent\textbf{Visual observation history mitigates effect of video-action segmentation errors.}
Errors by segmentation module lead to mis-classification of actions, in turn leading to erroneous action history.
To precisely study the effect of erroneous action history on \M's performance,
we perform controlled experiments wherein we add noise in the ground truth segmentation. This helps us tune desired segmentation error, irrespective of VideoCLIP's accuracy (in segmentation module). We achieve this by replacing a calculated \% of ground-truth actions by random actions. Decreasing performance, with increasing segmentation classification error, is expected and observed in  Fig.~\ref{fig:random noise}.
Further, we compare two models variations, VLaMP$(G,A_k,O_k)$ \ie the full model (uses both action and observation history) and VLaMP$(A,A_k)$ that does not have visual observation history.
As we increase segmentation classification error, the gap in the performance of these model variations increases.
We infer that visual observations add robustness in \T against video-action segmentation errors.
\begin{figure}
    \centering
    \includegraphics[width=\columnwidth]{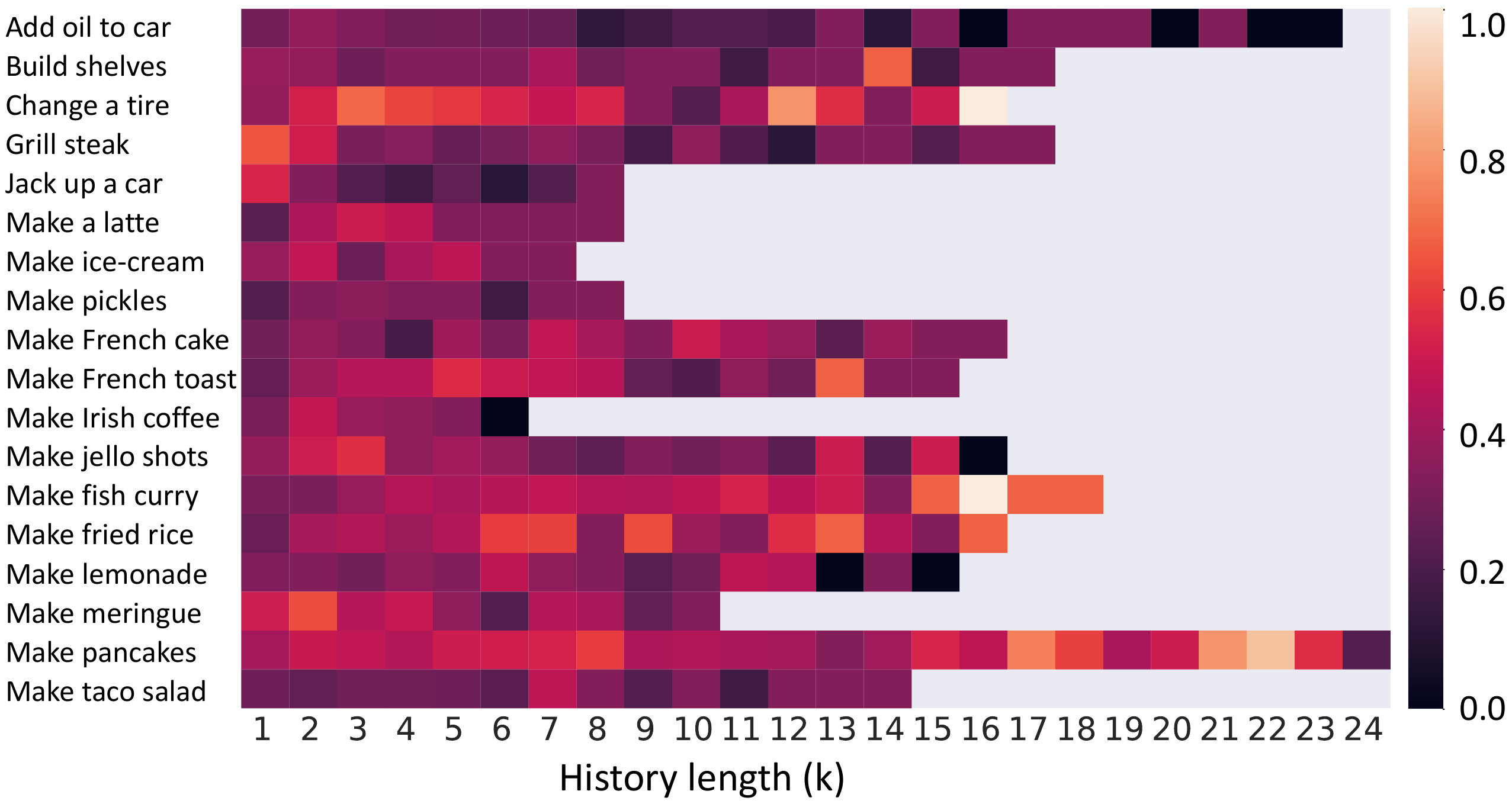}
    \vspace{-4mm}
    \caption{\textbf{Accuracy in the tail of long activities is challenging.} Longer history $(O_k, A_k)$ do not necessary lead to better accuracy. Further longer histories occur with increasing rarity, leading to learning challenges. (Plot shows mAcc for $l=3$ CrossTask \vs the \#steps in history $k$)
    }
    \label{fig:grid plot}
\end{figure}
%
%
%
%
%
%
%
%
%

%

%

\noindent\textbf{Errors do not drop as more history is provided.}
In \figref{fig:grid plot}, we show mAcc w.r.t. the number of steps $k$ in the history \ie length of $(O_k, A_k)$.
As more information is available (due to more steps in the history), one may expect the accuracy to improve.
However, this isn't the trend for many tasks. For \eg, note the accuracy drops for the task ``add oil to your car'' w.r.t $k$. Digging deeper, we find two reasons for this pattern.
\textit{First}, we find that tasks with long history like ``add oil to your car'' tend to have repetitive steps towards the end of their action trajectories, making it difficult to predict the precise number and the pattern of such repetitions in \T.
\textit{Second} is simply a data issue -- long trajectories are
also exponentially less frequent in the dataset. This forms the tail
of the data distribution of action sequences in the dataset
(see App.~\ref{app:more_quant} for statistics).
Consequently, prediction in this tail of activities, is a pertinent challenge for future research.

\FloatBarrier
\section{Conclusion}
\Task is a new and intuitive formulation to support planning from natural visual observations for assisting humans in day-to-day activities.
We benchmark \T using standardized suite of prior baselines and a new multi-modal sequence modeling formation of \M. The novel decomposition of a \M policy into  video action segmentation and forecasting leads to several efficiency and modeling benefits, that benefit the community, even beyond \T.
Particularly, this allows to leverage pre-trained LM that leads to significant performance gains.
Alternative decompositions, self-supervised pre-training objectives for \PTLM{s}, modeling multiple actors, and tapping into other modalities (audio, camera metadata, \etc)~\cite{MAE,mittal2022learning,liu2021cooperative,szot2023adaptive,g2p,zheng2023exif} are interesting ways forward for the community, on the exciting task of \T.
\vspace{-2mm}
\paragraph{Acknowledgements:}
The authors are indebted to \href{https://brian7685.github.io/}{Brian Chen} for his efforts to open-source the code and LTA experiments, and to Gideon Stocek for supporting with the cluster management. 
We thank Vasu Sharma, Tushar Nagarajan, and Homanga Bharadhwaj for their writing feedback. 
We thank Weichao Mao for early discussions and Asli Celikyilmaz for her constant guidance.

{\small
\bibliographystyle{ieee_fullname}
\bibliography{egbib}
}
\clearpage
\appendix
\section*{Appendix -- Pretrained Language Models as Visual Planners for Human Assistance}

We structure the supplementary material as follows:
\begin{itemize}\compresslist
    \item[~\ref{app:baselines}.] Implementation details of baselines: DDN~\cite{chang2020procedure}, GPT3-based~\cite{brown2020language, huang2022language} language-only method and most-probable actions.
    \item[~\ref{sec:inference}.] Step-by-step algorithm for inference, for reproducibility and technical details.
    \item[~\ref{app:training}.] Optimization, hardware, and training details associated with training \M.
    \item[~\ref{app:lta}.] Comparisons to Ego4D's LTA benchmark.
    \item[~\ref{app:more_quant}.] Expanded empirical results, benchmarking the above additional baselines, and deep-dive into error analysis.
\end{itemize}

\section{Baselines}
\label{app:baselines}
In this section we include more information about baselines that we benchmark on \T in experiments (\secref{sec:experiments}). \textit{First}, we include necessary details of reproducing the DDN~\cite{chang2020procedure} and how we keep it consistent and fair to the proposed \M. \textit{Second}, we provide two additional baselines -- a heuristic baseline, which leverages the structure of our goal-oriented activities for generating plans and a prompt-based baseline using a large LM. \textit{Finally}, we briefly discuss prior procedural planning methods, which we choose not to compare with.
As mentioned in the main paper, we also provide the std (standard error of mean) around the mean for various models in \tabref{tab:main-results-vlamp-ext}.

\subsection{DDN~\cite{chang2020procedure}}
\label{app:ddn_details}

\noindent\textbf{Technical Background.} Chang~\etal~\cite{chang2020procedure} proposed Dual Dynamics Network (DDN) for procedural planning.
The objective is to learn a latent space representation of observations and actions in addition to a dynamics and conjugate dynamics model that operate over this latent space. 
The latent representations and recurrent RNN-based dynamics model are learned together by minimizing a joint loss over predicted observations and actions. Such dynamics modeling in latent space is similar in spirit to the forecasting module in \M (\equref{eq:autoreg}). 

\begin{figure}[t]
    \begin{tabular}{c}
    \includegraphics[width=0.8\linewidth]{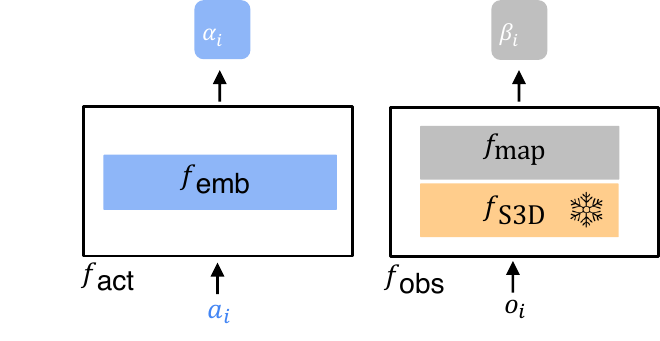}\vspace{-1mm}\\
    (a) Encoders\vspace{5mm}\\
    \includegraphics[width=0.7\linewidth]
    {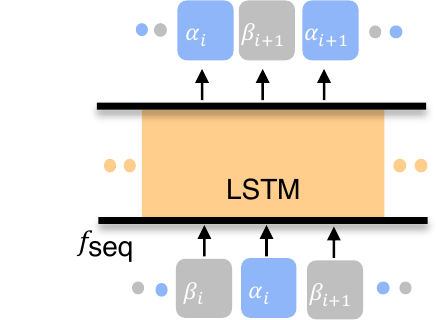}\\
    (b) Sequence Model\\
    \end{tabular}
    \caption{\textbf{DDN Implementation.} The method utilizes LSTM in the forecasting module for \T. Consistent to and analogous of \M's \figref{fig:forecasting model} in the main paper.}
    \label{fig:app:DDN}
\end{figure}

\noindent\textbf{Implementation.} As shown in \figref{fig:app:DDN}, we instantiate DDN for \T by using an LSTM-based~\cite{hochreiter1997long} $f_{\textrm{seq}}$ in the forecasting module,
which operates over the observation representations obtained using the same observation encoder $\VEnc$ consisting of pretrained S3D~\cite{s3d} and a mapper as \M and action representations from an embedding layer-based action encoder $\AEnc$. Unlike \M, where the mapper aims to project the visual observation representations into the input space of the pretrained LM, the mapper in DDN only provides trainable parameters to finetune the frozen S3D representations for downstream dynamics model. Both $f_{\textrm{seq}}$ and $\AEnc$ are initialized with random weights. Just as \M, DDN is trained using cross-entropy loss for predicted actions and mean-squared error for predicted observation representations to jointly learn $\VEnc$, $\AEnc$, and the sequence model.
At inference, the model is unrolled autoregressively  (with beam search as shown in Algo.~\ref{alg:inference-ext}), for prediction of both action and observation representations. These design choices are consistent with \M.

\begin{figure}[t]
    \centering
    \includegraphics[width=\columnwidth]{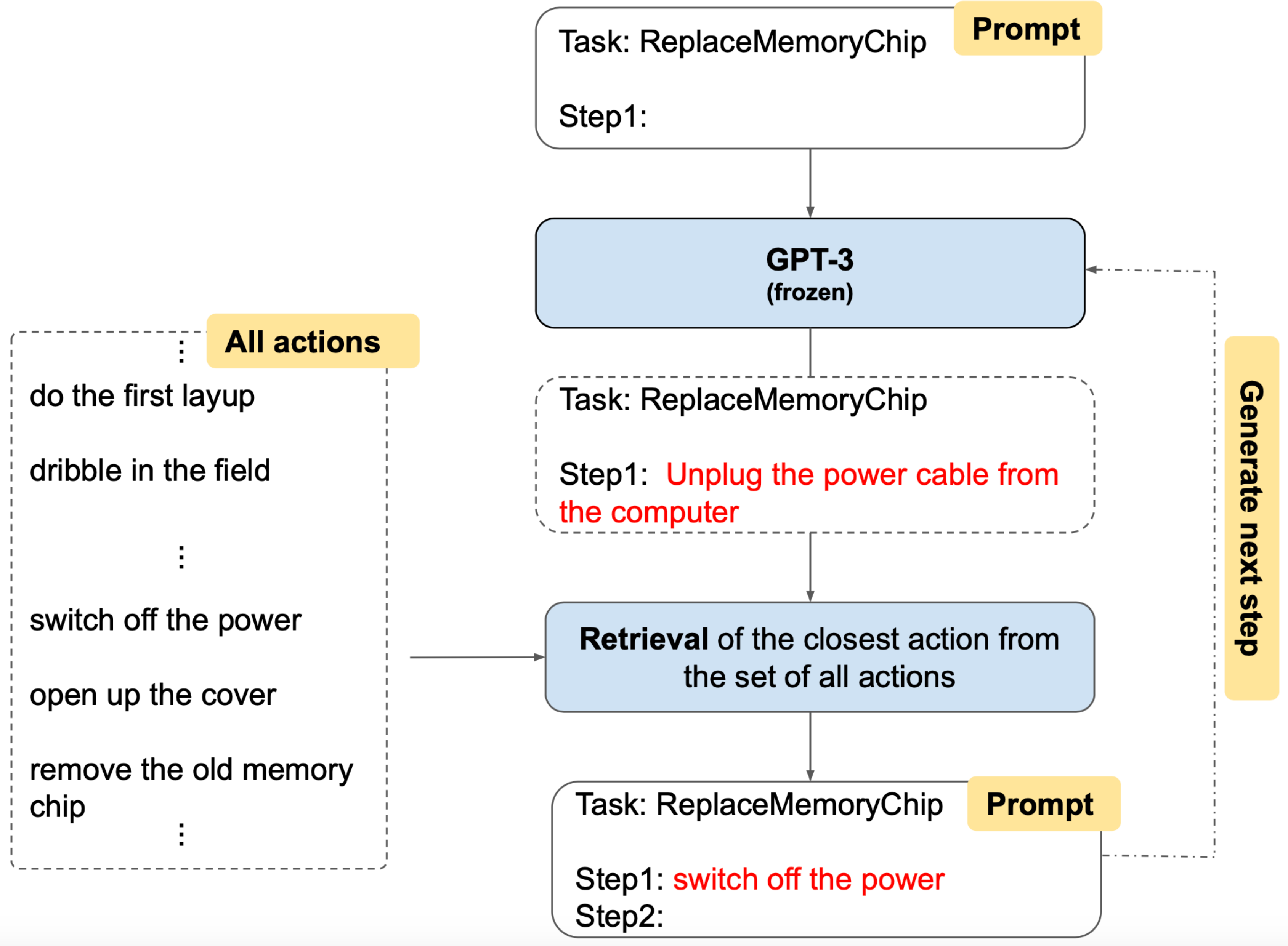}
    \caption{\textbf{GPT-3 as a planner based on~\cite{huang2022language}.} A \textit{language-only} baseline using GPT-3 on COIN. 
    GPT-3 is prompted autoregressively to generate next action based on the goal and the history of actions taken for the goal.}
    \label{fig:gpt3}
\end{figure}

\begin{algorithm*}[t]
\footnotesize
\caption{Inference for VLaMP and baselines}\label{alg:inference-ext}
\SetKwComment{Comment}{// }{ }
\KwData{encoded representations for history $H_k=(h_1, \dots, h_n)$,\,  beam size $B$}
\KwResult{plan for next $l$ steps $\hat{\mathcal T} = a_{k+1}, \cdots, a_{k+l}$}

$\mathcal H_0 \gets \{H_k\}$\Comment*[r]{Initialize the set of encoded trajectories with the history}
\For{$i = 1, \dots, l$ \Comment*[r]{predict actions for $l$ steps}}{ 
  
  $\mathcal H_i \gets \{H\diamond f_{\textrm{enc}}(a) \mid  H\in \mathcal H_{i-1},\, a\in \mathcal A\}$ \Comment*[r]{All single action extension at $i$-th step}
  $\Phi_i \gets \{\phi( H ) \mid H\in \mathcal H_i\}$\Comment*[r]{score each trajectory}
  \SetKwFunction{sorted}{sorted}
  \SetKwFunction{top}{top}
  
  $\tilde{\mathcal H}_i, \tilde{\Phi}_i \gets$ \top{B,\, \sorted{$\mathcal H_i, \Phi_i$}} \Comment*[r]{Keep $B$ highest scoring trajectories}
  
  $\tilde{\mathcal H}_{i}^0 \gets \tilde{\mathcal H}_i$\;
  \For{$u = 1, \dots, \delta$ \Comment*[r]{Predict $\delta$ observation representations autoregressively}}{
    $\tilde{\mathcal H}_i^u \gets \{H \diamond f_{\textrm{seq}}(H)\mid H\in \tilde{\mathcal H}_i^{u-1}\}$
  }
  $\mathcal H_i \gets \tilde{\mathcal H}_i^{\delta}$\; %
}
\SetKwFunction{readout}{readout}

$\hat{\mathcal T} \gets$ \readout{l, \top{1,\, \sorted{$\tilde{\mathcal H}_l, \tilde{\Phi}_l$}}} \Comment*[r]{Read out last $l$ actions from the top scoring beam}
\end{algorithm*}

\subsection{GPT-3 Planner} \label{app:additional baseline GPT3 planner}
Following Huang~\etal~\cite{huang2022language}, where the authors use a LLM as zero-shot planners, we too also experiment with prompting a frozen pretrained large language model (GPT-3) for \T. 
Specifically, a goal prompt and the current history of previously predicted actions (if available) are given as a prompt to the GPT-3 model~\cite{brown2020language}. 
Then the next actions for the given goal are generated autoregressively, consistent with other baselines like \M. 
As can be seen in Fig.~\ref{fig:gpt3}, this model has 2 stages: 1) next-action generation, and 2) action retrieval.
In the next-action generation stage, the model is given the prompt and generates the next action.
In the action retrieval stage, the generated next action is compared to all possible actions and the action that has the closest similarity to the generated action, is chosen and placed in the prompt. This step is required since we evaluate the generated plans by comparing with ground truth actions for the goal, where actions belong to a closed set as described in Sec.~\ref{subsec:evaluationofT}. We use \emph{text-davinci-003} backend for generation, and \emph{text-embedding-ada-002} for embedding, which is used in combination with cosine similarity to retrieve the closest action.
We perform our generation step zero-shot without giving any examples, as GPT3 can already follow the given prompt template, and we only generate one action at the time -- in an autoregressive manner. 

\begin{table*}[t]
\footnotesize
\centering
\begin{tabular}{@{}llccccccc@{}}
\toprule
\multirow{2}{*}{Dataset} &
  \multirow{2}{*}{Method} &
  $l=1$ &
  \multicolumn{3}{c}{$l=3$} &
  \multicolumn{3}{c}{$l=4$} \\ \cmidrule(l){3-9} 
 &
   &
  nAcc &
  SR &
  mAcc &
  mIOU &
  SR &
  mAcc &
  mIOU \\ \midrule
\multirow{4}{*}{CrossTask} &
  Random &
  0.9 {\normalsize ± 0.0} &
  0.0 {\normalsize ± 0.0} &
  0.9 {\normalsize ± 0.0} &
  1.5 {\normalsize ± 0.0} &
  0.0 {\normalsize ± 0.0} &
  0.9 {\normalsize ± 0.0} &
  1.9 {\normalsize ± 0.0} \\
 &
  Random w/ goal &
  13.2 {\normalsize ± 0.2} &
  0.3 {\normalsize ± 0.0} &
  13.4 {\normalsize ± 0.0} &
  23.6 {\normalsize ± 0.1} &
  0.0 {\normalsize ± 0.0} &
  12.7 {\normalsize ± 0.0} &
  27.8 {\normalsize ± 0.1} \\
 &
  DDN~\cite{chang2020procedure} &
  33.4 {\normalsize ± 0.5} &
  6.8 {\normalsize ± 0.3} &
  25.8 {\normalsize ± 0.5} &
  35.2 {\normalsize ± 0.6} &
  3.6 {\normalsize ± 0.2} &
  24.1 {\normalsize ± 0.4} &
  37.0 {\normalsize ± 0.4} \\
 &
  VLaMP \textit{(ours)} &
  \textbf{50.6 {\normalsize ± 1.4}} &
  \textbf{10.3 {\normalsize ± 0.4}} &
  \textbf{35.3 {\normalsize ± 1.1}} &
  \textbf{44.0 {\normalsize ± 1.0}} &
  \textbf{4.4 {\normalsize ± 0.2}} &
  \textbf{31.7 {\normalsize ± 1.0}} &
  \textbf{43.4 {\normalsize ± 0.9}} \\ \midrule
\multirow{4}{*}{COIN} &
  Random &
  0.1 {\normalsize ± 0.0} &
  0.0 {\normalsize ± 0.0} &
  0.1 {\normalsize ± 0.0} &
  0.2 {\normalsize ± 0.0} &
  0.0 {\normalsize ± 0.0} &
  0.1 {\normalsize ± 0.0} &
  0.2 {\normalsize ± 0.0} \\
 &
  Random w/ goal &
  24.5 {\normalsize ± 0.2} &
  1.7 {\normalsize ± 0.0} &
  21.4 {\normalsize ± 0.1} &
  42.7 {\normalsize ± 0.1} &
  0.3 {\normalsize ± 0.0} &
  20.1 {\normalsize ± 0.1} &
  47.7 {\normalsize ± 0.1} \\
 &
  DDN~\cite{chang2020procedure} &
  29.3 {\normalsize ± 0.3} &
  10.1 {\normalsize ± 0.4} &
  22.3 {\normalsize ± 0.4} &
  32.2 {\normalsize ± 0.6} &
  7.0 {\normalsize ± 0.3} &
  21.0 {\normalsize ± 0.4} &
  37.3 {\normalsize ± 0.3} \\
 &
  VLaMP \textit{(ours)} &
  \textbf{45.2 {\normalsize ± 0.8}} &
  \textbf{18.3 {\normalsize ± 0.1}} &
  \textbf{39.2 {\normalsize ± 0.3}} &
  \textbf{56.6 {\normalsize ± 0.5}} &
  \textbf{9.0 {\normalsize ± 0.3}} &
  \textbf{35.2 {\normalsize ± 0.2}} &
  \textbf{54.2 {\normalsize ± 0.5}} \\ \bottomrule
\end{tabular}%
\vspace{+1mm}
\caption{\textbf{Expanded version of \tabref{tab:main-results-vlamp}.} The mean $\pm$ ste.\ (standard error of mean) for various planning metrics obtained using 5 runs with different random seed are shown for \M and various baselines. Note: the models that don't change with random seeds are omitted. 
Note that the action history and observations are provided using the output of the action segmentation model and hence are noisy compared to the ground truth history.
}
\label{tab:main-results-vlamp-ext}
\end{table*}

\subsection{`Most Probable Action' baseline}\label{app:heuristic baselines}

In addition to the two intuitive, heuristic baselines, \emph{random} and \emph{random w/~goal}, we also use a stronger heuristic baseline called `most probable action' baseline. 
Here we describe this baseline in detail. 
The idea is to leverage the fact that procedural activities are \textit{highly structured}, \ie, certain actions occur together or occur in a certain order. 
We bake this into a simple model with Markov assumption, that the probability distribution of the next action $a_{k+1}$ given the current action $a_{k}$ to predict future actions.
Akin to random w/ goal baseline, we also evaluate a goal-conditioned most probable action baseline, that uses a goal-specific set of actions $\mathcal A_G\subset \mathcal A$ during sampling. 
Since these most probable baselines, provide a probability distribution over the actions, we can employ beam search (for fairness, with the same beam size same as \M) and pick the highest scoring plan. 
%

%
%
%
%
%

\subsection{On Porting More Baselines}
Next, we briefly include some procedural planning approaches and reasons why they cannot be directly leveraged for rigorous and fair evaluation. Wherever possible, we include our best attempts to compare with them.
\begin{table*}[t]
\centering
\resizebox{\linewidth}{!}{%
\begin{tabular}{@{}cccccccccc@{}}
\toprule
 &
  \begin{tabular}[c]{@{}c@{}}beam \\ size ($B$)\end{tabular} &
  \begin{tabular}[c]{@{}c@{}}per node \\ beam size ($b$)\end{tabular} &
  GPU &
  \begin{tabular}[c]{@{}c@{}}GPU\\ memory\end{tabular} &
  \begin{tabular}[c]{@{}c@{}}Num \\ GPUs\\ (inference)\end{tabular} &
  \begin{tabular}[c]{@{}c@{}}Num\\ GPUs\\ (training)\end{tabular} &
  \begin{tabular}[c]{@{}c@{}}Avg. time\\ (training)\end{tabular} &
  \begin{tabular}[c]{@{}c@{}}Avg. time\\ (inference)\end{tabular} &
  \begin{tabular}[c]{@{}c@{}}batch\\ size\\ (training)\end{tabular} \\ \midrule
CrossTask &
  10 &
  3 &
  NVIDIA  A100 &
  80GB &
  1 GPU/model &
  1 GPU/model &
  2 s/batch &
  7.4 s/example &
  4 \\
COIN &
  3 &
  3 &
  NVIDIA  A100 &
  80GB &
  3 GPUs/model &
  1 GPU/model &
  2 s/batch &
  6.1 s/example &
  4 \\ \bottomrule
\end{tabular}%
}
\caption{Hyperparameters and compute information for \M.}
\label{tab:hyperparameters}
\end{table*}
\noindent\textbf{PlaTe}~\cite{sun2022plate}: This is similar to DDN, albeit with a Transformer~\cite{vaswani2017attention} as the sequence model instead of an LSTM~\cite{hochreiter1997long}. However, unlike DDN (and \M), PlaTe uses separate Transformer-based models for state and action prediction. We adopt an approach that allowed us to tap into this while being consistent and fair in evaluation. Therefore, instead of directly adapting PlaTe for \T as we did with DDN, we provide an ablation on \M, which uses a Transformer trained from scratch as the sequence model (row \textbf{R} in \tabref{tab:ablations-short}). 
    
\noindent\textbf{P3IV}~\cite{zhao2022p3iv}: This employs a significantly different modeling framework compared to DDN and PlaTe. 
Specifically, P3IV leverages a memory-augmented transformer as the sequence model and a probabilistic generative model to capture the noise and variability in predicted sequences. 
The authors report significant performance gain on the task of procedure planning, over DDN and PlaTe.
However, P3IV relies on the visual observation of the goal already completed, even at inference time. This is necessary to condition their generative model towards encoding multiple plans from start to goal. Since P3IV needs the observations of goal completed, it is incompatible to the motivation and the very premise of \T. 

\begin{figure*}[t]
    \centering
%
    \centering
    \subfigure[$\mathcal A_G$ across COIN and CrossTask]{\label{fig:app:actions-per-task}
        \includegraphics[width=0.25\linewidth]{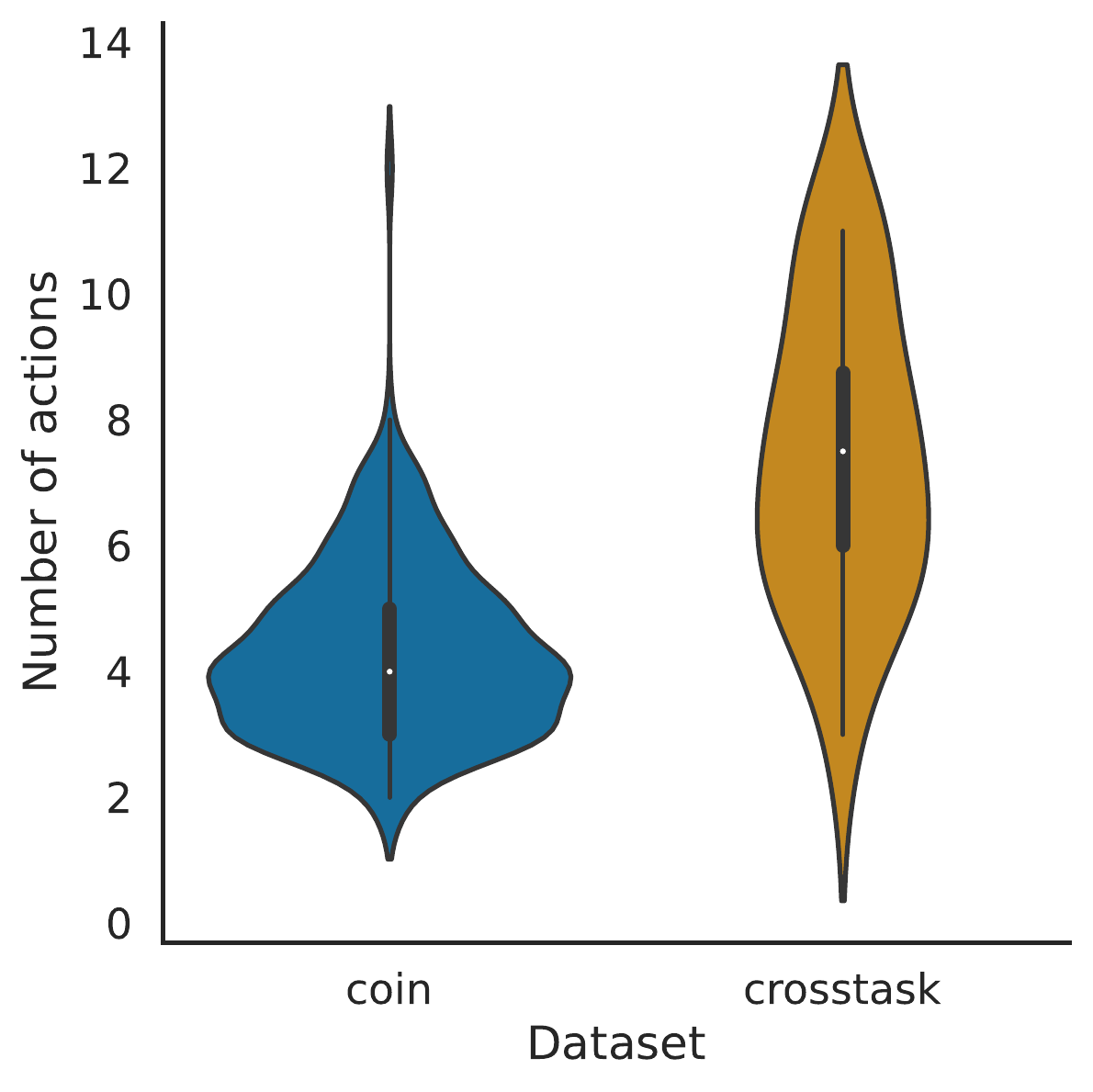}}
    \hfill
    \subfigure[Mean accuracy (mAcc) averaged across goals]{\label{fig:app:mean mAcc}
        \includegraphics[width=0.35\linewidth]{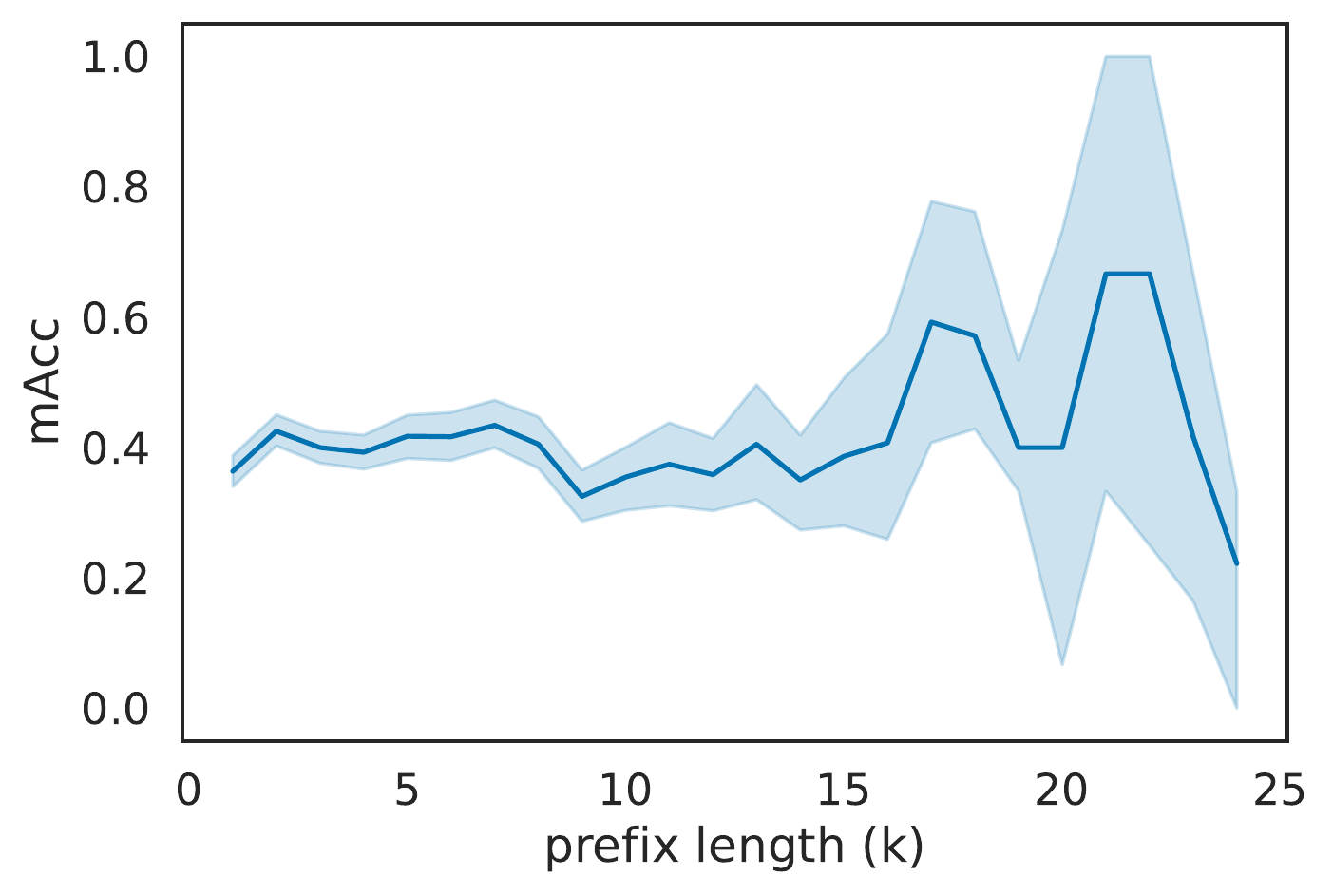}}
    \hfill
    \subfigure[Number of samples for each prefix length (length of history)]{\label{fig:app:count vs k}\includegraphics[width=0.35\linewidth]{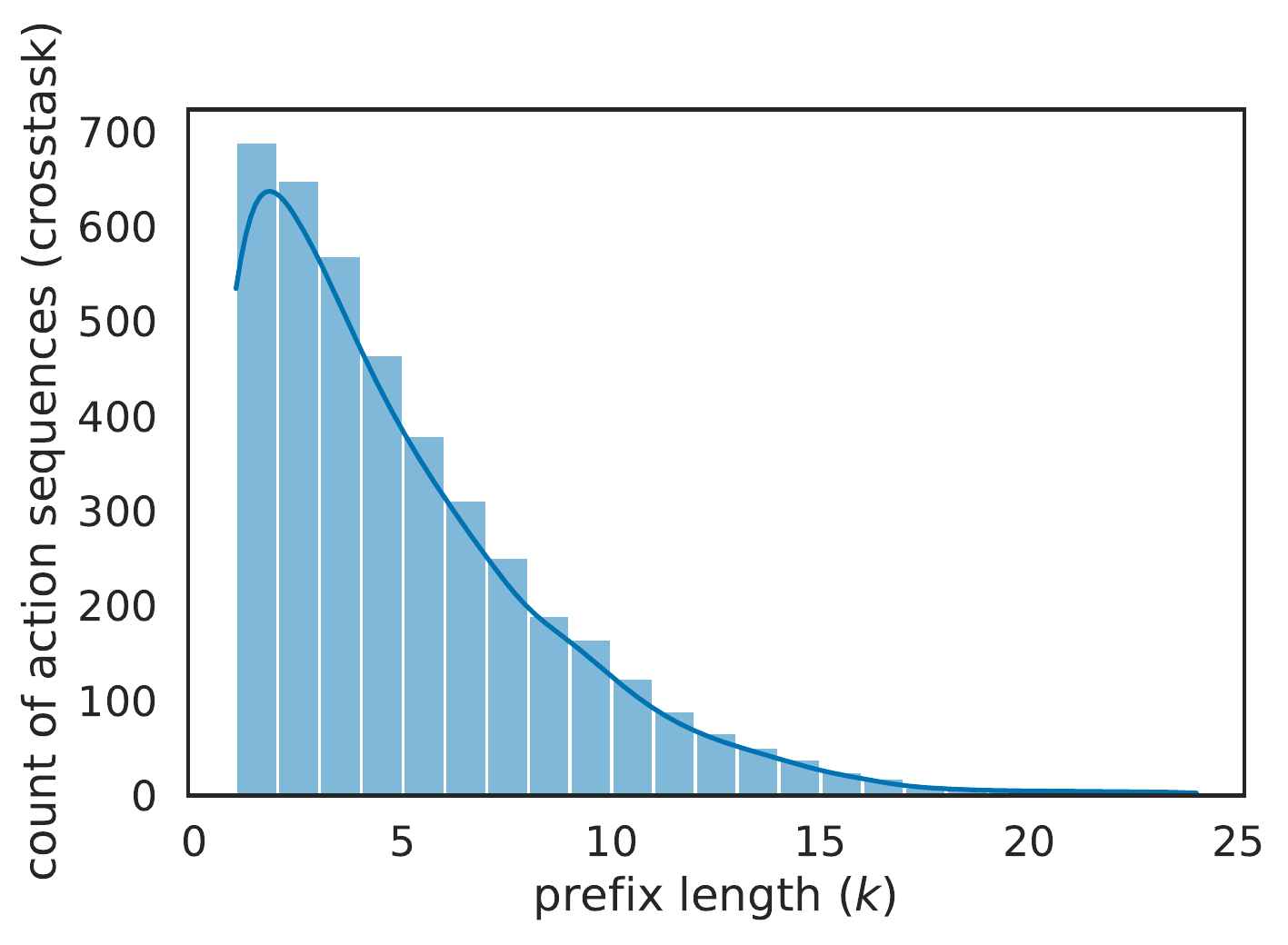}}
    \label{fig:summed metrics}
    \caption{\textbf{Zooming into the tails.} (a) Average size of goal-specific action set $\mathcal A_G$ across COIN and CrossTask datasets. COIN has a relatively smaller mean than CrossTask, which reduces the difficulty of \T on COIN. (b) Mean accuracy (mAcc) \vs the history length $k$ of various goals from CrossTask. Interestingly, plan generation for goals with longer history is difficult and prone to higher errors as reflected in the mAcc (reasoning included in \secref{app:more_quant}). 
    (c) Longer sequences are also less frequent in the dataset.
    This contributes to high variance in performance for such sequences.}
    \label{fig:app:performance w.r.t k}
\end{figure*}

\section{Inference for \T (\algref{alg:inference-ext})}\label{app:training/opt/inference}
\label{sec:inference}
In order to predict a sequence of next actions, we run the sequence model, autoregressively to predict both the action and observation tokens, with beam search on the action sequence.
The inference algorithm is detailed in Algo.~\ref{alg:inference-ext}.
We first encode the history into a sequence of representations $H_k$ as described in \secref{sec:forecasting model}, and initialize our set of encoder trajectories $\mathcal H_0$ using this single representation trajectory (line 1 in \algref{alg:inference-ext}).
Then we start the inference procedure that runs for $l$ steps (line 2).
At each step $i$ we first infer the next action and then also predict the representations for the observation that follows it.
To do the former, each representation trajectory in $\mathcal H_i$ is extended with the representations of each action in the action set $\mathcal A$ (line 4).
At this point, if, for instance, $\mathcal H_{i-1}$ had $n$ trajectories, then after line 4, $\mathcal H_i$ will have $n\times |\mathcal A|$ trajectories.
This is a temporary blow-up--at line 6, we score all $n\times |\mathcal A|$ trajectories and keep only top $B$ trajectories.
Here, to balance diversity, we keep no more than $b$ trajectories with exactly same history.
The parameter $b$ is usually referred to as \emph{per node beam size}.
Once we have $B$ trajectories in $\mathcal H_i$, we auto-regressively predict the next $\delta$ tokens corresponding to the next observation, thus completing one out of the $l$ steps of inference.
This process it repeated $l$ times to generate a plan consisting of $l$ actions.
We make this process efficient by storing the hidden state of the transformer and limiting the forward pass only on the new representations at each step. This is a common practice for transformer based models in NLP. 
Due to beam search, the inference process is slower than training as shown in \tabref{tab:hyperparameters}.

\section{\M Training (\tabref{tab:hyperparameters})}
\label{app:training}
Unlike inference where a video with $K$ steps results into $K-4$ examples, during training, like with language model pre-training, we use a single forward pass to compute loss for all tokens.
Moreover, inference also uses beam search making it more memory intensive.
Thus, the training is much faster and cheaper as compared to the inference. 
The details of the compute used for each training and inference run is shown in \tabref{tab:hyperparameters}.

\section{Comparison to Ego4D LTA Benchmark}
\label{app:lta}
In prior work, Ego4D's Long-term Action Anticipation~\cite{grauman2022ego4d} benchmark task is also highly relevant to \T. Hence, we dedicate a discussion of similarities and contrasts. We hope this helps the reader accurately place these two tasks in our community's diverse research goals and directions. 

Consistent to \T, LTA also focuses on predicting a sequence of future actions given prior visual context for free-form human interaction. Unlike LTA, \T specifically entails \textit{goal-oriented} activities and indeed a natural language goal prompt is of key importance to the definition of \T. So while the forecasting suite in Ego4D aspires to understand human motion, we are instead keen to create assistive agents that can interact and assist humans in their tasks. 

Since LTA does not allow access or model the user's goal, recent approaches for LTA including the winning model for Ego4D 2022 LTA challenge -- ICVAE~\cite{Mascaro_2023_WACV} have to go via an additional step of inferring the intention of the user. This provides more impetus to our goal-conditioned and human-assitive design choice and motivation for \T.
This is empirically backed as well, as we show in ablation (row 1 in Tab.~\ref{tab:ablations-short}) -- goal-conditioning is crucial for \T. 

We also evaluated \M on Ego4D's LTA benchmark. Both the VideoCLIP-based segmentation module and the forecasting module in \M were finetuned on LTA's train split. Table~\ref{tab:ego4d-vlamp} shows the performance of \M on the validation split of LTA for $l=20$ future actions, following the LTA task definition. The extremely long future horizon of prediction makes beam search impractical. Hence, we don't perform beam search with \M for this experiment. \M outperforms the best performing LTA baseline, which uses a visual encoder followed by a transformer-based aggregator~\cite{grauman2022ego4d}. While Ego4D used Slowfast encoder for the LTA baseline, we perform this experiment using S3D encoder for a fair comparison with \M.

\begin{table}[]
\centering
\resizebox{0.9\columnwidth}{!}{%
\begin{tabular}{@{}ccccc@{}}
\cmidrule(l){3-5}
\multicolumn{1}{l}{} & \multicolumn{1}{l}{} & \multicolumn{3}{c}{ED@(l=20)}                   \\ \midrule
Model                & Encoder              & Verb          & Noun           & Action         \\ \midrule
VLaMP (ours)         & S3D                  & \textbf{0.73} & \textbf{0.772} & \textbf{0.932} \\
LTA Baseline~\cite{grauman2022ego4d}         & S3D                  & 0.745         & 0.779          & 0.941          \\ \bottomrule
\end{tabular}%
}
\caption{\textbf{\M on Ego4D LTA's validation split.} Edit distance for verb, noun, action prediction for future 20 steps is shown (\textbf{lower} the better).}
\label{tab:ego4d-vlamp}
\end{table}

\begin{figure*}[t]
    \centering
    \includegraphics[width=0.8\linewidth]{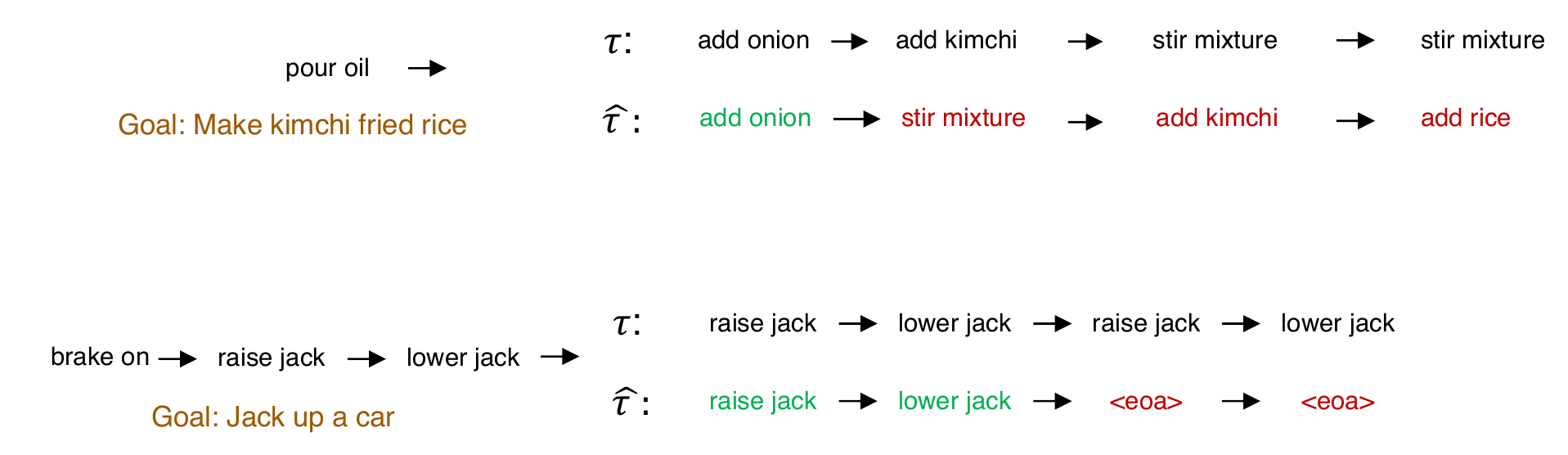}
    \caption{\textbf{Qualitative Error Analysis for \T.} Ground truth plan $\mathcal T$ and the predicted plan $\hat {\mathcal T}$ by \M for the goal prompt of ``making kimchi fried rice'' (\textit{top}) and `` Jack up the car" (\textit{bottom}).
    Errors made by \M can be attributed to \textit{repetitions in actions}. Details are included in \secref{app:more_quant}. Briefly, (1) uncertainty in the number of times actions are repeated and (2) existence of equivalent plans for achieving the same goal, are contribute heavily to the errors for \T. 
    In the top, note the action `stir mixture' is repeated consecutively in the ground truth, but the model predicts it only once. 
    Moreover, both the ground truth and the predicted plans have correct steps for adding kimchi and onion but their \textit{order is different}. Similar repetitions result into errors for the goal of jacking up the car.}
    \label{fig:qual}
\end{figure*}

\begin{figure}[t]
    \centering
    \includegraphics[width=\columnwidth]{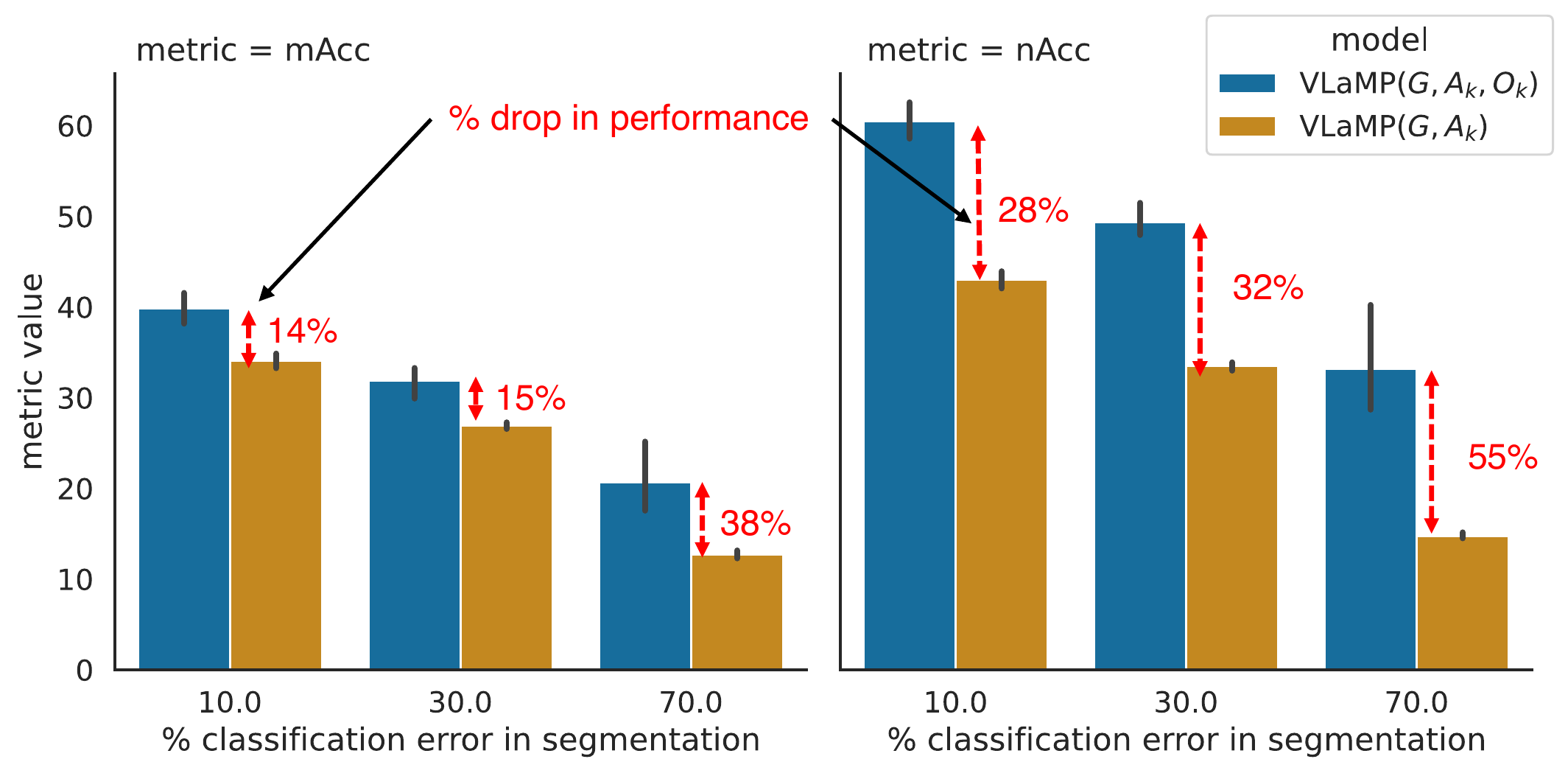}
    \caption{\textbf{Effect of segmentation errors.} The figure zooms in on two metrics mAcc and nAcc from Figure \ref{fig:random noise}.  As the classification error in segmentation, which is shown along the x-axis increases, the performance gap between the model with access to observation history \M($G, A_k, O_k$) and that with access only to the action history \M($G, A_k$) increases.}
    \label{fig:app:effect of segmentation errors}
\end{figure}

\section{Additional Quantitative Results}
\label{app:more_quant}

\noindent\textbf{Most probable action baseline.} As shown in Table~\ref{tab:main-results-vlamp}, the performance of the heuristic baselines -- \emph{most probable action w\slash~ goal}, and \emph{Random w\slash~goal} (\ie the baselines with actions restricted to the set of actions seen with the corresponding goal) is quite high for COIN dataset.
We find that this is due to the relatively small cardinality of the action set for goals in COIN, i.e. the average size of $\mathcal A_G$ for different $G$s as discussed below. 

\noindent\textbf{Action distribution analysis.} We dig deeper into the above finding in \figref{fig:app:actions-per-task}. Particularly, we plot the distribution of number of actions $\left|\mathcal A_G\right|$ w.r.t $G$ and find them to be quite different in COIN vs. CrossTask. Here, $\left|\mathcal A_G\right| \in \mathcal A$ represents the set of goal-specific actions from the larger set of actions $\mathcal A$ for each dataset. Specifically, the average size of $\mathcal A_G$ is $4.3$ and $7.3$, respectively for COIN and CrossTask, reducing the difficulty of \T in COIN. Also, notice the long-tails in the distribution of CrossTask, making it even more challenging. 

\noindent\textbf{Zooming into the tails and higher errors.} In \figref{fig:app:mean mAcc}, we plot the number of steps in the history ($k$) \vs the mean accuracy (mAcc), averaged across all goals in CrossTask on \T.
We find that plan generation with \textit{longer history leads to higher errors} as well as higher variance in performance. We believe this trend emerges due to \textit{two reasons}. 

\textit{First}, the presence of repetitive steps in certain goals is high in longer history. Moreover, we find that longer the history the wider is the space of possible plans (intuitively, multiple modes exist in the plan distribution landscape), which lead to higher variance. We illustrate this in \figref{fig:qual} with a qualitative result, for the example goal of `making kimchi fried rice', the action `stir mixture' repeatedly occurs between various actions involving the addition of ingredients like onion, kimchi, rice, etc. However, the number of times \emph{stir mixture} occurs varies sporadically. 
For instance, for the ground truth plan in the first example in \figref{fig:qual}, the `stir mixture' is missing between `add onion' and `add kimchi', but occurs twice after `add kimchi', before adding other ingredients.
Due to this sporadic variability, the predicted plan gets IoU of 75\% on this example, but mAcc and SR of 25\% and 0, respectively.
Another common source of errors is repetition of sub-sequences of actions depending on the visual signal in the ground truth.
Specifically, as seen in the second example in \figref{fig:qual}, which shows an action trajectory for the goal of `jack up a car', the sub-sequence (`raise jack', `lower jack'), is repeated three times.
In this example, the repetition is due to overshooting the target height of the raised car. However, for a planning model, that only sees the visual input till $k=3$ or time $t_{k}$, it is not possible to guess whether the car will overshoot (undershoot, respectively) the target height after the next application of `raise jack' (`lower jack', respectively).

\textit{Second}, as analysed in \figref{fig:app:count vs k}, \textit{longer trajectories are exponentially less frequent} in the dataset -- forming the tail of the data distribution of action sequences in the dataset. This also contributes to high variance in performance for such sequences.

\noindent\textbf{Segmentation errors.}
As we note in \secref{sec:error analysis} and \figref{fig:random noise} of the main paper, segmentation errors are detrimental for \T. As the mis-classification error in the segmentation model increases, the difference in the performance of \M ($G, A_k, O_k$), i.e. the model with access to observation history, and \M ($G, A_k$), the model working only on the action history, increases. A detailed version of \figref{fig:random noise} is included in \figref{fig:app:effect of segmentation errors} with a focus on mean accuracy (mAcc) and next-step accuracy (nAcc)\footnote{Refer to the definitions of metrics in \secref{subsec:evaluationofT}}. 
Moreover, since the observation history has higher influence on predicting the immediate next action as discussed in Sec.~\ref{sec:error analysis}, the performance drop due to segmentation classification error is higher in nAcc as compared to mAcc.

\end{document}